\documentclass{article} %
\usepackage{iclr2027_conference,times}

\usepackage{amsmath,amsfonts,bm}

\def\eqref#1{equation~\ref{#1}}

\def\1{\bm{1}}

\DeclareMathAlphabet{\mathsfit}{\encodingdefault}{\sfdefault}{m}{sl}
\SetMathAlphabet{\mathsfit}{bold}{\encodingdefault}{\sfdefault}{bx}{n}

\usepackage{hyperref}
\usepackage{url}

\title{\framework: Procedural Shielding of Tool-Using LLM Agents}

\author{Ashwani Anand$^{1,}$\thanks{Equal contribution.}\phantom{$^{*}$}, Ivi Chatzi$^{1,}$\footnotemark[1]\phantom{$^{*}$}, Ritam Raha$^{2}$ \& Anne-Kathrin Schmuck$^{1}$ \\
$^{1}$Max Planck Institute for Software Systems, Kaiserslautern, Germany \\
$^{2}$TNO-ESI, Eindhoven, The Netherlands \\
\texttt{\{ashwani,ichatzi,akschmuck\}@mpi-sws.org}, \texttt{ritam.raha@tno.nl}
}

\usepackage[disable]{todonotes}
\usepackage{xspace}
\usepackage{amsmath,amssymb,amsthm}
\usepackage{stmaryrd}
\usepackage{enumitem}
\usepackage{hyperref}
\usepackage{tabularx}
\usepackage{longtable}
\usepackage{microtype}
\usepackage{parskip}
\usepackage{wrapfig}
\usepackage{newfile}
\usepackage{csquotes}

\usepackage[normalem]{ulem}

\usepackage{multirow}
\usepackage{fancyvrb}
\usepackage{fvextra}   %
\usepackage{textcomp}
\DeclareUnicodeCharacter{2192}{\textrightarrow}   %
\protected\def\notesemdash{\ifdim\fontdimen3\font=0pt --\else\textemdash\fi}
\DeclareUnicodeCharacter{2014}{\notesemdash}
\DefineVerbatimEnvironment{notebox}{Verbatim}{breaklines,breaksymbolleft={},fontsize=\footnotesize,frame=single,rulecolor=\color{gray},framesep=4pt}
\usepackage{pifont}
\usepackage{subcaption}

\usepackage{array}
\usepackage{booktabs}

\usepackage{xcolor}
\usepackage{colortbl}
\usepackage{tikz}
\usetikzlibrary{positioning, arrows.meta, fit, calc, backgrounds, shapes, shadows}

\usepackage{setup/testcasebox}

\newif\ifshowcomments
\showcommentstrue
\showcommentsfalse   %
\ifshowcomments
  \usepackage{colorblind}
  \usepackage{mention}
\fi

\usepackage[section]{placeins} %
\usepackage{float} %
\usepackage{needspace} %

\ifshowcomments
\else
\fi

\definecolor{opscol}{HTML}{1F4FA0}
\definecolor{medcol}{HTML}{0E7C5C}
\definecolor{auditcol}{HTML}{B83838}
\definecolor{factcol}{HTML}{555555}

\newcommand{\framework}{\textsc{MENTOR}\xspace}
\newcommand{\chal}[1]{\ensuremath{\langle}C#1\ensuremath{\rangle}\xspace}   %
\newcommand{\mantra}{\textsc{MANTRA}\xspace}
\newcommand{\taubench}{\ensuremath{\tau^2}-bench\xspace}

\newcommand{\ltlform}[1]{\ensuremath{#1}}
\newcommand{\ltlG}{\ensuremath{\mathbf{G}}}
\newcommand{\ltlF}{\ensuremath{\mathbf{F}}}
\newcommand{\ltlX}{\ensuremath{\mathbf{X}}}
\newcommand{\ltlO}{\ensuremath{\mathbf{O}}}
\newcommand{\ltlU}{\ensuremath{\mathbin{\mathbf{U}}}}
\newcommand{\ltlW}{\ensuremath{\mathbin{\mathbf{W}}}}
\newcommand{\ltland}{\ensuremath{\wedge}}
\newcommand{\ltlor}{\ensuremath{\vee}}
\newcommand{\ltlnot}{\ensuremath{\neg}}
\newcommand{\ltlimplies}{\ensuremath{\rightarrow}}
\newcommand{\ltlprop}[1]{\ensuremath{\mathsf{#1}}}

\newcommand{\doc}{\mathcal{D}\xspace}

\newcommand{\tool}{\mathcal{T}\xspace}

\newcommand{\code}[1]{\texttt{#1}}

\newcommand{\ckcall}{\textbf{\texttt{CALL}}}
\newcommand{\cknocall}{\textbf{\texttt{NO-CALL}}}
\newcommand{\ckor}{\textbf{\texttt{OR}}}
\newcommand{\ckbefore}{\textbf{\texttt{BEFORE}}}
\newcommand{\ckafter}{\textbf{\texttt{AFTER}}}
\newcommand{\ckfollows}{\textbf{\texttt{FOLLOWS}}}
\newcommand{\ckprecedes}{\textbf{\texttt{PRECEDES}}}

\newcolumntype{L}[1]{>{\raggedright\arraybackslash}p{#1}}

\newcommand{\passAtOne}{\textbf{Pass@}$\mathbf{1}$}
\newcommand{\passAtK}{\textbf{Pass@}$\mathbf{k}$}
\newcommand{\passAtFive}{\textbf{Pass@}$\mathbf{5}$}
\newcommand{\passSupK}{$\mathbf{Pass\text{\textasciicircum}k}$}
\newcommand{\passSupFive}{$\mathbf{Pass\text{\textasciicircum}5}$}

\newcommand{\passSupThree}{$\mathbf{Pass\text{\textasciicircum}3}$}

\definecolor{cmarkgreen}{RGB}{0,140,60}
\definecolor{xmarkred}{RGB}{200,30,30}
\definecolor{pmarkorange}{RGB}{220,140,0}

\iclrfinalcopy %
\begin{document}

\label{beginningofdocument}
\newoutputstream{docstatus}
\openoutputfile{main.ds}{docstatus}
\addtostream{docstatus}{Preparing}
\closeoutputstream{docstatus}

\maketitle
\lhead{Preprint}

\begin{abstract}

Tool-using LLM agents are increasingly used to automate workflows governed by procedural manuals, such as company rules for booking travel or procuring hardware. These manuals are written in natural language for humans and describe general requirements over such complex workflows. Ensuring compliance of LLM agents with these manuals is difficult, and state-of-the-art techniques use either LLM judges or manually written rules to forbid problematic tool-calls locally at runtime. This is insufficient because (i) only blocking tool calls keeps the agent clueless about how to make correct progress towards fulfilling its task, and (ii) can lead to overly conservative behavior of the agent, as tool calls might only be problematic in certain contexts.
To overcome these challenges, we introduce \framework, which (i) additionally provides guidance for the agent to make procedural-compliant progress on the task, and (ii) ensures that this compliance is context-aware. We achieve these goals by building a novel history-dependent runtime shield that is obtained from the procedural manual via a multi-step autoformalization pipeline. At its core, \framework utilizes a game-based technique to ensure robustness against runtime uncertainty without over-resticting agents' behavior.
This yields a formally assisted framework for procedural shielding of LLM agents that improves compliance well beyond existing methods. In particular, \framework lifts the
compliance of shielded light-weight open-source LLMs above that of non-shielded proprietary models. Since a shield is synthesized once per manual and then serves every task, a \framework-shielded agent becomes a viable choice for repeatedly executed workflows, potentially reducing deployment costs when the offline construction cost is amortized over repeated tasks.

\end{abstract}

\section{Introduction}\label{sec:introduction}

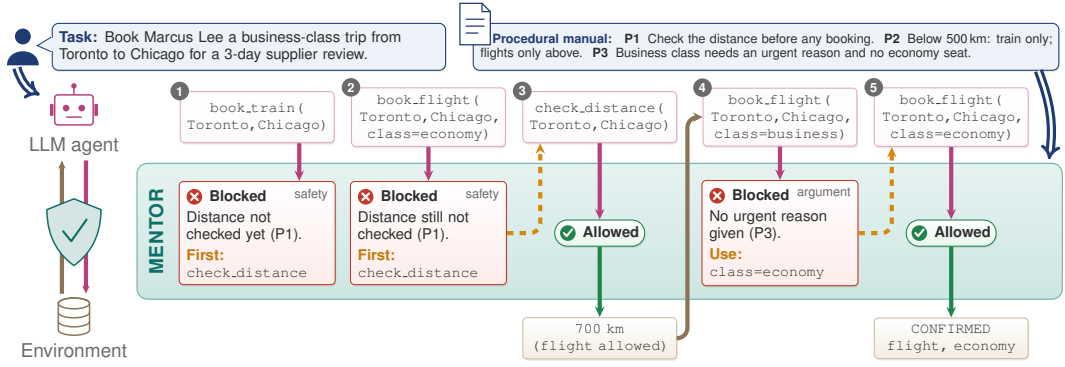
\begin{figure*}[t]
  \vspace{-1cm}
\centering\resizebox{\linewidth}{!}{%
\begin{tikzpicture}[x=1cm,y=1cm,font=\sffamily\scriptsize,
  /utils/exec={%
    \definecolor{shTop}{HTML}{E6F3F1}%
    \definecolor{shBand}{HTML}{CDE6E2}%
    \definecolor{shEdge}{HTML}{7FB8B3}%
    \definecolor{shInk}{HTML}{0F6E66}%
    \definecolor{agentC}{HTML}{B8407A}%
    \definecolor{agentEdge}{HTML}{E3A8C4}%
    \definecolor{sand}{HTML}{8A7455}%
    \definecolor{sandEdge}{HTML}{CDBFA6}%
    \definecolor{inputC}{HTML}{1F3B73}%
    \definecolor{inputEdge}{HTML}{9FB3D1}%
    \definecolor{inputFill}{HTML}{EEF2F9}%
    \definecolor{blockFill}{HTML}{FDECEA}%
    \definecolor{blockEdge}{HTML}{C0392B}%
    \definecolor{hintFill}{HTML}{FFF4DE}%
    \definecolor{hintEdge}{HTML}{D68910}%
    \definecolor{okFill}{HTML}{E8F6EC}%
    \definecolor{okEdge}{HTML}{1E8449}%
    \definecolor{toolFill}{HTML}{F3EEE4}%
    \definecolor{toolEdge}{HTML}{CDBFA6}%
    \definecolor{ink}{HTML}{2E2E2E}%
    \definecolor{muted}{HTML}{6B6B6B}%
    \def\markcross{\tikz[baseline=-0.55ex,sharp corners]{\fill[blockEdge] (0,0) circle (0.13);
      \draw[white,line width=1pt,line cap=round] (-0.055,-0.055) -- (0.055,0.055) (-0.055,0.055) -- (0.055,-0.055);}}%
    \def\marktick{\tikz[baseline=-0.55ex,sharp corners]{\fill[okEdge] (0,0) circle (0.13);
      \draw[white,line width=1pt,line cap=round,line join=round] (-0.065,0.0) -- (-0.02,-0.05) -- (0.07,0.06);}}%
  },
  card/.style={draw=#1,shade,shading=axis,top color=white,bottom color=#1!6!white,line width=0.5pt,rounded corners=2pt,
    inner sep=3pt,text=ink,align=center,
    drop shadow={opacity=0.18,shadow xshift=0.5pt,shadow yshift=-0.9pt}},
  call/.style={card=agentEdge,font=\ttfamily\scriptsize,minimum height=0.85cm,text width=2.35cm},
  verdict/.style={card=#1,text width=2.35cm,align=left,inner sep=3.5pt},
  block/.style={verdict=blockEdge,top color=white,bottom color=blockFill},
  ok/.style={card=okEdge,top color=white,bottom color=okFill,inner sep=2.5pt,rounded corners=6pt},
  result/.style={card=toolEdge,top color=white,bottom color=toolFill,text width=2.35cm,font=\ttfamily\scriptsize},
  input/.style={draw=inputC,fill=inputFill,line width=0.5pt,rounded corners=3pt,inner sep=3.5pt,text=ink,align=left,
    drop shadow={opacity=0.18,shadow xshift=0.5pt,shadow yshift=-0.9pt}},
  kind/.style={anchor=north east,inner sep=3.5pt,text=muted,font=\sffamily\tiny},
  lane/.style={text=muted,font=\sffamily\footnotesize,align=center},
  loopl/.style={text=muted,font=\sffamily\tiny},
  stepnum/.style={circle,fill=muted,text=white,font=\sffamily\bfseries\tiny,inner sep=1.2pt},
  flow/.style={-{Stealth[length=4pt,width=3.5pt]},line width=0.7pt},
  inflow/.style={-{Stealth[length=4pt,width=3.5pt]},line width=0.7pt,draw=inputC},
  outflow/.style={-{Stealth[length=5pt,width=4.5pt]},line width=2pt,draw=sand,rounded corners=5pt},
  callin/.style={-{Stealth[length=5pt,width=4.5pt]},line width=2pt,draw=agentC},
  fwdout/.style={-{Stealth[length=5pt,width=4.5pt]},line width=2pt,draw=okEdge},
  adopt/.style={-{Stealth[length=5pt,width=4.5pt]},line width=2pt,draw=hintEdge,dashed,rounded corners=5pt},
  pics/robot/.style={code={%
    \draw[fill=white,draw=#1,line width=0.7pt,rounded corners=1.5pt] (-0.32,-0.25) rectangle (0.32,0.25);
    \fill[#1] (-0.13,0.03) circle (0.055) (0.13,0.03) circle (0.055);
    \draw[#1,line width=0.7pt,line cap=round] (-0.12,-0.12) -- (0.12,-0.12);
    \draw[#1,line width=0.7pt] (0,0.25) -- (0,0.38); \fill[#1] (0,0.40) circle (0.045);
    \fill[#1,rounded corners=0.5pt] (-0.40,-0.10) rectangle (-0.34,0.10) (0.34,-0.10) rectangle (0.40,0.10);}},
  pics/shield/.style={code={%
    \path[fill=#1!18,draw=#1,line width=0.7pt,line join=round]
      (0,0.36) -- (0.30,0.24) .. controls (0.30,-0.10) and (0.14,-0.28) .. (0,-0.38)
      .. controls (-0.14,-0.28) and (-0.30,-0.10) .. (-0.30,0.24) -- cycle;
    \draw[#1,line width=1pt,line cap=round,line join=round] (-0.12,0.00) -- (-0.03,-0.11) -- (0.14,0.13);}},
  pics/gear/.style={code={%
    \foreach \a in {0,45,...,315}{\fill[#1,rotate=\a] (-0.07,0.22) rectangle (0.07,0.34);}
    \fill[#1] (0,0) circle (0.26); \fill[white] (0,0) circle (0.10);}},
  pics/doc/.style={code={%
    \path[fill=white,draw=#1,line width=0.6pt,line join=round]
      (-0.22,-0.30) -- (-0.22,0.30) -- (0.10,0.30) -- (0.22,0.18) -- (0.22,-0.30) -- cycle;
    \draw[#1,line width=0.6pt] (0.10,0.30) -- (0.10,0.18) -- (0.22,0.18);
    \draw[#1,line width=0.6pt,line cap=round] (-0.13,0.07) -- (0.13,0.07) (-0.13,-0.03) -- (0.13,-0.03) (-0.13,-0.13) -- (0.04,-0.13);}},
  pics/person/.style={code={%
    \fill[#1] (0,0.16) circle (0.11);
    \path[fill=#1] (-0.22,-0.24) .. controls (-0.22,0.02) and (0.22,0.02) .. (0.22,-0.24) -- cycle;}},
  pics/db/.style={code={%
    \path[fill=white,draw=#1,line width=0.7pt] (-0.26,0.22) -- (-0.26,-0.22) arc (180:360:0.26 and 0.09) -- (0.26,0.22) arc (0:180:0.26 and 0.09);
    \draw[#1,line width=0.7pt] (0,0.22) ellipse (0.26 and 0.09);
    \draw[#1,line width=0.7pt] (-0.26,0.07) arc (180:360:0.26 and 0.09) (-0.26,-0.08) arc (180:360:0.26 and 0.09);}},
]
\def\xa{4.20}\def\xb{7.05}\def\xc{9.90}\def\xd{12.90}\def\xe{15.75}   %
\def\yI{5.60}   %
\def\yA{4.43}   %
\def\yS{2.50}   %
\def\yT{0.75}   %
\def\yBt{3.66}  %
\def\yBb{1.40}  %
\def\xBl{2.20}  %

\pic[scale=1.2] at (0.30,\yI-0.02) {person=inputC};
\node[input,anchor=west,font=\sffamily\scriptsize,text width=6.3cm,minimum height=0.78cm] (task) at (0.78,\yI)
  {\textcolor{inputC}{\textbf{Task:}} Book Marcus Lee a business-class trip from Toronto to Chicago for a 3-day supplier review.};
\path[fill=inputFill,draw=inputC,line width=0.5pt,sharp corners]
  ($(task.west)+(-0.01,0.16)$) -- ($(task.west)+(-0.24,0.02)$) -- ($(task.west)+(-0.01,-0.08)$);
\path[fill=inputFill] ($(task.west)+(-0.02,-0.06)$) rectangle ($(task.west)+(0.03,0.14)$);
\node[input,anchor=east,font=\sffamily\tiny,text width=9.5cm,minimum height=0.78cm] (policy) at (17.55,\yI)
  {\hspace*{0.2cm}\textcolor{inputC}{\textbf{Procedural manual:}}\quad
   \textbf{P1}\; Check the distance before any booking.\;
   \textbf{P2}\; Below 500\,km: train only; flights only above.\;
   \textbf{P3}\; Business class needs an urgent reason and no economy seat.};
\pic[scale=1.15] at ($(policy.north west)+(0.04,-0.02)$) {doc=inputC};
\draw[inputC,double,double distance=1.4pt,line width=1.5pt,-{Implies}]
  (policy.south -| 17.20,0) to[bend left=14] (17.38,\yBt+0.07);
  \draw[inputC,double,double distance=1.4pt,line width=1.5pt,-{Implies}]
  (0.20,\yI-0.35) to[bend right=24] (0.58,\yA+0.37);

\pic at (1.15,\yA+0.12) {robot=agentC};
\node[lane] at (1.15,\yA-0.48) {LLM agent};

\pic at (1.15,\yT+0.36) {db=sand};
\node[lane] at (1.15,\yT-0.20) {Environment};

\begin{scope}[on background layer]
  \path[shade,shading=axis,top color=shTop,bottom color=shBand,rounded corners=5pt] (\xBl,\yBb) rectangle (17.60,\yBt);
  \draw[shEdge,line width=0.6pt,rounded corners=5pt] (\xBl,\yBb) rectangle (17.60,\yBt);
\end{scope}
\node[text=shInk,font=\sffamily\footnotesize\bfseries,inner sep=0pt,rotate=90] at (\xBl+0.32,{(\yBt+\yBb)/2}) {\framework};
\draw[flow,line width=2pt,draw=agentC] (1.35,\yA-0.72) -- (1.35,\yT+0.74);
\draw[flow,line width=2pt,draw=sand] (0.95,\yT+0.74) -- (0.95,\yA-0.72);
\pic[scale=1.5] at (1.15,{(\yBt+\yBb)/2}) {shield=shInk};

\node[call] (a1) at (\xa,\yA) {book\_train(\\Toronto,Chicago)};
\node[stepnum] at (a1.north west) {1};
\node[block] (s1) at (\xa,\yS)
  {\markcross\hspace{0.12cm}\textbf{Blocked}\\[2pt]
   Distance not checked yet (P1).\\[2pt]
   \textcolor{hintEdge!90!black}{\textbf{First:}} \texttt{check\_distance}};
\node[kind] at (s1.north east) {safety};
\draw[callin] (a1.south -| 4.90,0) -- (s1.north -| 4.90,0);

\node[call] (a2) at (\xb,\yA) {book\_flight(\\Toronto,Chicago,\\class=economy)};
\node[stepnum] at (a2.north west) {2};
\node[block] (s2) at (\xb,\yS)
  {\markcross\hspace{0.12cm}\textbf{Blocked}\\[2pt]
   Distance still not checked (P1).\\[2pt]
   \textcolor{hintEdge!90!black}{\textbf{First:}} \texttt{check\_distance}};
\node[kind] at (s2.north east) {safety};
\draw[callin] (a2) -- (s2);

\node[call] (a3) at (\xc,\yA) {check\_distance(\\Toronto,Chicago)};
\node[stepnum] at (a3.north west) {3};
\node[ok] (s3) at (\xc,\yS) {\marktick\hspace{0.12cm}\textbf{Allowed}};
\node[result] (t3) at (\xc,\yT) {700 km\\(flight allowed)};
\draw[callin] (a3) -- (s3);
\draw[fwdout] (s3) -- (t3);
\draw[adopt] (s2.east) -| ($(a3.south west)+(0.30,0)$);

\node[call] (a4) at (\xd,\yA) {book\_flight(\\Toronto,Chicago,\\class=business)};
\node[stepnum] at (a4.north west) {4};
\draw[outflow,-{Stealth[length=4.5pt,width=4.5pt]}] (t3.east) -| (11.36,\yA) [rounded corners=1.5pt] -- (a4.west);
\node[block] (s4) at (\xd,\yS)
  {\markcross\hspace{0.12cm}\textbf{Blocked}\\[2pt]
   No urgent reason given (P3).\\[2pt]
   \textcolor{hintEdge!90!black}{\textbf{Use:}} \texttt{class=economy}};
\node[kind] at (s4.north east) {argument};
\draw[callin] (a4) -- (s4);

\node[call] (a5) at (\xe,\yA) {book\_flight(\\Toronto,Chicago,\\class=economy)};
\node[stepnum] at (a5.north west) {5};
\node[ok] (s5) at (\xe,\yS) {\marktick\hspace{0.12cm}\textbf{Allowed}};
\node[result] (t5) at (\xe,\yT) {CONFIRMED\\flight, economy};
\draw[callin] (a5) -- (s5);
\draw[fwdout] (s5) -- (t5);
\draw[adopt] (s4.east) -| ($(a5.south west)+(0.30,0)$);

\end{tikzpicture}}
\vspace{-0.5cm}
\caption{\textbf{Running example.} Our procedural shield \framework is generated from a procedural manual (top right) and applied to a travel-booking task (top left). \framework shields the interaction between an LLM agent (pink) and the environment (brown). \framework blocks tool calls (red), allows tool calls (green), and provides guidance towards task completion (orange).} %
\label{fig:shield-example}
\end{figure*}

The abilities of tool-using large language model (LLM) agents allow them to be used to automate workflows such as hardware procurement~\citep{zdravkovic2022ai,ozman2025systematic}, patient diagnosis~\citep{tu2025towards,ghezloo2025pathfinder} or travel bookings~\citep{dwivedi2025artificial}. Such workflows are governed by \emph{procedural manuals}, written in natural language for human operators, that specify \emph{how} such tasks must be executed. For example, Fig.~\ref{fig:shield-example} (top right) shows a simple manual for booking a business trip. %

While such workflows may seem natural for humans, LLM agents have been observed to struggle with them, both in existing work about evaluating compliance post-hoc~\citep{li2025sopbench,nandi2026sop,anand2026mantra}, and in our own experiments. For instance in the domain of Fig.~\ref{fig:shield-example}, checking agents' tool call traces  reveals they often attempt to book business class after inventing a fake ``urgent reason'', or never book anything at all. %

To aid LLM agents in such tasks, we introduce \framework, a procedural
shield that checks proposed tool calls against formalized requirements
and guides agents towards fulfilling outstanding obligations.
Shielding is a well-established technique for policies obtained by reinforcement learning (RL), where a shield monitors the actions a policy proposes at runtime and blocks those that violate a formal specification~\citep{konighofer2022correct,Review23SafetyFilterHsuHuFisac}.
We develop this approach for tool-using LLM agents around four
interconnected challenges.

\textit{\chal{1} Context-dependence \& guidance.} A procedural manual, as e.g.\ the one in Fig.~\ref{fig:shield-example}, induces a partial temporal order over tool calls, so whether a tool call complies is \emph{context-dependent}  (e.g., booking a flight is allowed in step~5 but not in step~2).
This context-dependence introduces \emph{obligations} an agent needs to fulfill before another necessary call would be allowed (e.g., checking the distance before any booking) and it must eventually make some specific tool calls to complete its task (e.g., the booking itself).
Blocking a call without identifying a compliant next step can leave
the agent repeatedly attempting blocked calls and failing to complete
its task, as illustrated by our experiments
(Fig.~\ref{fig:stalling} in App.~\ref{app:failure-traces-baselines}).
Shields that also incorporate context-dependent progress guidance were only recently introduced for RL as \emph{strategy template based runtime shields (\textsc{STARS})}~\citep{anand2026follow}.
\framework adapts this idea to LLM agents, by first auto-formalizing the  manual into Linear Temporal Logic (LTL) and then synthesizing a strategy template from it.

\textit{\chal{2} Argument-based constraints.} Context-dependence does not only affect the order of tool calls, but also their arguments. For example, booking an economy class flight is allowed without an urgent reason in step~5, whereas booking business class is not in step~4.
The values these arguments can take often range over unbounded domains, such as strings, 
and might only be revealed at runtime, making it challenging to express constraints over them. 
The importance of checking argument values is illustrated by our
analysis of SOP-Bench~\citep{nandi2026sop}, whose evaluator accepts some
incorrect traces because it does not check the
relevant tool arguments (App.~\ref{app:sopbench}).
\framework addresses this problem with argument constraints using Satisfiability Modulo Theories (SMT) that are checked and, if violated, repaired at runtime.

\textit{\chal{3} Natural language.} Any formally assisted framework for procedural shielding of LLM agents needs to translate natural language texts within the manual, the task description, and tool outputs into formal artifacts.
In our work, we automatically translate the manual into a formal specification offline.
Then, at runtime, we observe that both cheap proprietary LLMs and fine-tuned small open-weights LLMs can effectively ground the task description and tool outputs. %

\textit{\chal{4} Delivering the shield.} A classical shield for an RL policy intervenes directly on the action the policy proposes. %
In the context of LLM agents, however, actions are tool calls which are written token by token.
Depending on the availability of the agent's next-token distribution, \framework can deliver the blocking and advice of tool calls to the agent in two ways:
(i) via a prompt shield that injects textual feedback in the tool call outputs,
and (ii) via a token shield, which masks the tokens of unsafe calls and boosts those of progressing calls during generation.

To show its impact on procedural compliance of shielded agents, we evaluate \framework on 3 benchmarks covering 8 domains using both proprietary and open-weights agents.
The experiments show that \framework clearly outperforms unshielded agents and other guarding methods, and lifts the compliance of small open-weight models above those of unshielded proprietary ones.
We provide all our code and parameters at \href{https://anonymous.4open.science/r/mentor-llm-shield/}{https://anonymous.4open.science/r/mentor-llm-shield/}.

\smallskip
\noindent\textbf{Related Work.}
Shielding LLM agents has received increasing attention very recently, but none of the existing approaches addresses the outlined challenges appropriately. We compare them here and in Table~\ref{tab:shield-comparison} in App.~\ref{app:related-work} in more detail.

One line of work on LLM shielding techniques delegate compliance decisions to a guardrail LLM~\citep{xiang2024guardagent,hoover2026dynaguard,kang2026policyguide} which can interpret natural language (\chal{3}) within the manual, tasks, and tool outputs.
However, the resulting judgements can be unreliable due to the stochastic nature of LLMs. 
Alternatively, deterministic guards hard-code procedural constraints, which are either written by hand in a dedicated language~\citep{wang2025agentspec,winston2026solver,palumbo2026formal,reddy2026reason,kamath2025enforcing} or extracted by an LLM~\citep{shi2025progent,mondl2026autoformalization}. Such deterministic guards, however, cannot interpret natural language tool outputs during their execution.
As a result, they cannot provide rich enough context-dependent guidance to the agent (\chal{1}).
In addition, manually translating the procedural manual into formal constraints can make shield synthesis labor intensive and hard to extend to new domains, and this task is known to be difficult even for experts~\citep{czepa2019modeling,greenman2023little}.
\framework combines the strengths of both approaches by using a refinement-based auto-formalization technique to build a deterministic tool-call evaluator which is paired with a labeler to ground tool call outputs.

\begin{figure*}[t]
\centering
\begin{tikzpicture}[remember picture,x=1cm,y=1cm,font=\sffamily\fontsize{6.2}{7.2}\selectfont,
  /utils/exec={%
    \definecolor{ink}{HTML}{1F2933}%
    \definecolor{muted}{HTML}{5B6773}%
    \definecolor{slate}{HTML}{3B4C6B}%
    \definecolor{slateEdge}{HTML}{B4BFCF}%
    \definecolor{agentC}{HTML}{B8407A}%
    \definecolor{agentEdge}{HTML}{E3A8C4}%
    \definecolor{inputC}{HTML}{1F3B73}%
    \definecolor{inputFill}{HTML}{EEF2F9}%
    \definecolor{teal}{HTML}{0F6E66}%
    \definecolor{tealEdge}{HTML}{7FB8B3}%
    \definecolor{tealTop}{HTML}{E6F3F1}%
    \definecolor{tealBot}{HTML}{CDE6E2}%
    \definecolor{sand}{HTML}{8A7455}%
    \definecolor{sandEdge}{HTML}{CDBFA6}%
    \definecolor{sandFill}{HTML}{F3EEE4}%
    \definecolor{okC}{HTML}{2F9E63}%
    \definecolor{hintC}{HTML}{D98E1F}%
    \definecolor{blockC}{HTML}{C4453A}%
    \definecolor{ltlC}{HTML}{6B4FA0}%
    \definecolor{smtC}{HTML}{2E7BA6}%
    \definecolor{okFill}{HTML}{E8F6EC}%
    \definecolor{blockFill}{HTML}{FDECEA}%
    \definecolor{ltlEdge}{HTML}{B9A9D6}%
    \definecolor{smtEdge}{HTML}{A4C6DB}%
    \def\markcross{\tikz[baseline=-0.55ex,sharp corners]{\fill[blockC] (0,0) circle (0.11);
      \draw[white,line width=0.9pt,line cap=round] (-0.045,-0.045) -- (0.045,0.045) (-0.045,0.045) -- (0.045,-0.045);}}%
    \def\marktick{\tikz[baseline=-0.55ex,sharp corners]{\fill[okC] (0,0) circle (0.11);
      \draw[white,line width=0.9pt,line cap=round,line join=round] (-0.055,0.0) -- (-0.015,-0.045) -- (0.06,0.05);}}%
  },
  card/.style={draw=#1,shade,shading=axis,top color=white,bottom color=#1!6!white,
    line width=0.5pt,rounded corners=2.5pt,inner sep=3pt,text=ink,align=center,
    drop shadow={opacity=0.15,shadow xshift=0.4pt,shadow yshift=-0.9pt}},
  input/.style={draw=inputC,fill=inputFill,line width=0.5pt,rounded corners=3pt,inner sep=3pt,text=ink,align=left,
    drop shadow={opacity=0.15,shadow xshift=0.4pt,shadow yshift=-0.9pt}},
  chk/.style={card=tealEdge,minimum height=0.68cm,minimum width=2.1cm,inner sep=2pt},
  thinbox/.style={draw=#1,fill=white,line width=0.5pt,rounded corners=2.5pt,inner sep=0pt},
  panel/.style={draw=#1,shade,shading=axis,top color=white,bottom color=#1!35!white,rounded corners=6pt,line width=0.6pt},
  ptitle/.style={anchor=north west,font=\sffamily\fontsize{7.2}{8.5}\selectfont\bfseries,inner sep=4pt},
  phase/.style={anchor=north west,text=slate,font=\sffamily\fontsize{6}{7}\selectfont\bfseries,inner sep=3.5pt},
  btitle/.style={font=\sffamily\fontsize{6.8}{8}\selectfont,text=ink,inner sep=2pt},
  lab/.style={text=muted,font=\sffamily\fontsize{5.2}{6}\selectfont,inner sep=1.5pt},
  lane/.style={text=muted,font=\sffamily\fontsize{6.2}{7.2}\selectfont,align=center,inner sep=1pt},
  tag/.style={fill=#1,text=white,font=\sffamily\fontsize{4.6}{5}\selectfont\bfseries,inner sep=1.3pt,rounded corners=1pt},
  flow/.style={-{Stealth[length=4pt,width=3.5pt]},line width=0.7pt,draw=muted,rounded corners=4pt},
  inflow/.style={double,double distance=1.3pt,dotted,line width=1.5pt,-{Implies},draw=inputC},
  callf/.style={-{Stealth[length=5pt,width=4.5pt]},line width=2pt,draw=agentC,rounded corners=5pt},
  fwd/.style={-{Stealth[length=5pt,width=4.5pt]},line width=2pt,draw=okC,rounded corners=5pt},
  outp/.style={-{Stealth[length=5pt,width=4.5pt]},line width=2pt,draw=sand,rounded corners=5pt},
  outthin/.style={-{Stealth[length=4pt,width=3.5pt]},line width=0.7pt,draw=sand,rounded corners=4pt},
  schema/.style={-{Stealth[length=4pt,width=3.5pt]},line width=0.6pt,draw=sand,dashed,rounded corners=4pt},
  advice/.style={-{Stealth[length=5pt,width=4.5pt]},line width=2pt,draw=hintC,dashed,rounded corners=5pt},
  advthin/.style={-{Stealth[length=4pt,width=3.5pt]},line width=0.9pt,draw=hintC,dashed,rounded corners=4pt},
  chkcard/.style={card=#1,text width=1.7cm,minimum height=1.25cm,inner sep=3pt,font=\sffamily\fontsize{5.8}{7}\selectfont},
  okpill/.style={card=okC,top color=white,bottom color=okFill,inner sep=2.2pt,rounded corners=6pt,font=\sffamily\fontsize{5.8}{7}\selectfont\bfseries},
  blockpill/.style={card=blockC,top color=white,bottom color=blockFill,inner sep=2.2pt,rounded corners=6pt,font=\sffamily\fontsize{5.8}{7}\selectfont\bfseries},
  kite/.style={diamond,aspect=2.2,draw=tealEdge,shade,shading=axis,top color=white,bottom color=tealEdge!6!white,line width=0.5pt,inner sep=0.5pt,align=center,text=ink,font=\sffamily\fontsize{5.2}{6}\selectfont,drop shadow={opacity=0.15,shadow xshift=0.4pt,shadow yshift=-0.9pt}},
  pics/robot/.style={code={%
    \draw[fill=white,draw=#1,line width=0.7pt,rounded corners=1.5pt] (-0.32,-0.25) rectangle (0.32,0.25);
    \fill[#1] (-0.13,0.03) circle (0.055) (0.13,0.03) circle (0.055);
    \draw[#1,line width=0.7pt,line cap=round] (-0.12,-0.12) -- (0.12,-0.12);
    \draw[#1,line width=0.7pt] (0,0.25) -- (0,0.38); \fill[#1] (0,0.40) circle (0.045);
    \fill[#1,rounded corners=0.5pt] (-0.40,-0.10) rectangle (-0.34,0.10) (0.34,-0.10) rectangle (0.40,0.10);}},
  pics/shield/.style={code={%
    \path[fill=#1!18,draw=#1,line width=0.7pt,line join=round]
      (0,0.36) -- (0.30,0.24) .. controls (0.30,-0.10) and (0.14,-0.28) .. (0,-0.38)
      .. controls (-0.14,-0.28) and (-0.30,-0.10) .. (-0.30,0.24) -- cycle;
    \draw[#1,line width=1pt,line cap=round,line join=round] (-0.12,0.00) -- (-0.03,-0.11) -- (0.14,0.13);}},
  pics/gear/.style={code={%
    \foreach \a in {0,45,...,315}{\fill[#1,rotate=\a] (-0.07,0.22) rectangle (0.07,0.34);}
    \fill[#1] (0,0) circle (0.26); \fill[white] (0,0) circle (0.10);}},
  pics/doc/.style={code={%
    \path[fill=white,draw=#1,line width=0.6pt,line join=round]
      (-0.22,-0.30) -- (-0.22,0.30) -- (0.10,0.30) -- (0.22,0.18) -- (0.22,-0.30) -- cycle;
    \draw[#1,line width=0.6pt] (0.10,0.30) -- (0.10,0.18) -- (0.22,0.18);
    \draw[#1,line width=0.6pt,line cap=round] (-0.13,0.07) -- (0.13,0.07) (-0.13,-0.03) -- (0.13,-0.03) (-0.13,-0.13) -- (0.04,-0.13);}},
  pics/db/.style={code={%
    \path[fill=white,draw=#1,line width=0.7pt] (-0.26,0.22) -- (-0.26,-0.22) arc (180:360:0.26 and 0.09) -- (0.26,0.22) arc (0:180:0.26 and 0.09);
    \draw[#1,line width=0.7pt] (0,0.22) ellipse (0.26 and 0.09);
    \draw[#1,line width=0.7pt] (-0.26,0.07) arc (180:360:0.26 and 0.09) (-0.26,-0.08) arc (180:360:0.26 and 0.09);}},
  pics/person/.style={code={%
    \fill[#1] (0,0.16) circle (0.11);
    \path[fill=#1] (-0.22,-0.24) .. controls (-0.22,0.02) and (0.22,0.02) .. (0.22,-0.24) -- cycle;}},
  pics/graph/.style={code={%
    \draw[#1,line width=0.6pt] (-0.45,0.20) -- (-0.05,0.0) -- (0.35,0.22) (-0.05,0.0) -- (-0.05,-0.30) (0.35,0.22) -- (0.35,-0.25);
    \fill[white,draw=#1,line width=0.6pt] (-0.45,0.20) circle (0.09) (0.35,0.22) circle (0.09) (-0.05,-0.30) circle (0.09);
    \fill[white,draw=#1,line width=0.6pt] (-0.14,-0.09) rectangle (0.04,0.09) (0.26,-0.34) rectangle (0.44,-0.16);}},
]
\begin{scope}[on background layer]
  \path[fill=black!4,draw=slateEdge,line width=0.6pt,rounded corners=6pt] (4.35,4.70) rectangle (14.20,6.15);
  \path[fill=black!4,draw=slateEdge,line width=0.6pt,rounded corners=6pt] (0.28,0.95) rectangle (4.15,6.15);
\end{scope}
\node[phase] at (4.35,6.15) {Build \framework (offline)};
\node[phase] at (0.30,6.15) {Use \framework (online)};

\pic[scale=1.1] at (0.40,6.73) {person=inputC};
\node[input,anchor=west,minimum height=0.62cm,text width=2.95cm] (task) at (0.80,6.75)
  {\textcolor{inputC}{\textbf{Task}}\\natural-language request};
\path[fill=inputFill,draw=inputC,line width=0.5pt,sharp corners]
  ($(task.west)+(-0.01,0.14)$) -- ($(task.west)+(-0.20,0.02)$) -- ($(task.west)+(-0.01,-0.07)$);
\path[fill=inputFill] ($(task.west)+(-0.02,-0.05)$) rectangle ($(task.west)+(0.03,0.12)$);
\node[input,anchor=east,minimum height=0.62cm] (man) at (14.15,6.75)
  {\hspace*{0.15cm}\textcolor{inputC}{\textbf{Procedural manual} $\mathcal{D}$}\\\hspace*{0.15cm}natural-language rules};
\pic[scale=0.95] at ($(man.north west)+(0.04,-0.02)$) {doc=inputC};
\node[input,draw=sand,fill=sandFill,anchor=east,minimum height=0.62cm] (ts) at (11.30,6.75)
  {\hspace*{0.15cm}\textcolor{sand}{\textbf{Tool schema $\tool$}}\\\hspace*{0.15cm}structured tool descriptions};
\pic[scale=0.7] at ($(ts.north west)+(0.06,-0.04)$) {gear=sand};
\draw[inflow] (3.50,6.35) -- (3.50,5.95);
\draw[inflow,sand] (10.50,6.35) -- (10.50,5.95);
\draw[inflow] (12.90,6.35) -- (12.90,5.95);

\node[thinbox=agentEdge,minimum width=1.5cm,minimum height=0.95cm] (ag) at (1.80,5.30) {};
\pic[scale=0.9] at (1.80,5.37) {robot=agentC};
\node[lane] at (1.80,4.97) {LLM agent};

\node[card=slateEdge,btitle,minimum height=0.6cm,anchor=center,inner sep=2.5pt] (af) at (12.60,5.25)
  {\tikz[baseline=-0.6ex]\pic[scale=0.5]{robot=slate};\hspace{0.08cm}Autoformalization};
\pic[scale=1.2] at (8.80,5.67) {doc=ltlC};  \node[tag=ltlC] at (9.05,5.40) {LTL};
\pic[scale=1.2] at (9.85,5.25) {doc=smtC};  \node[tag=smtC] at (10.10,4.98) {SMT};
\draw[flow] (af.west) -- (10.55,5.25) -- (10.55,5.67) -- (9.12,5.67);
\draw[flow] (af.west) -- (10.17,5.25);
\node[card=slateEdge,btitle,minimum height=0.6cm,anchor=center,inner sep=2.5pt] (tc) at (6.10,5.25)
  {\tikz[baseline=-0.6ex]\pic[scale=0.7]{graph=slate};\hspace{0.08cm}Template computation};
\draw[flow] (8.48,5.67) -- (8.05,5.67) -- (8.05,5.25) -- (tc.east);

\begin{scope}[on background layer]
  \path[shade,shading=axis,top color=tealTop,bottom color=tealBot,rounded corners=6pt] (0.45,2.55) rectangle (3.80,3.80);
  \draw[tealEdge,line width=0.6pt,rounded corners=6pt] (0.45,2.55) rectangle (3.80,3.80);
\end{scope}
\pic[scale=0.7] at (0.45,3.80) {shield=teal};
\node[text=teal,font=\sffamily\fontsize{7.2}{8.5}\selectfont\bfseries,inner sep=1pt,anchor=south] at (2.17,2.57) {\framework};
\node[card=tealEdge,btitle,minimum height=0.5cm,anchor=center,inner sep=2.5pt] (ds) at (3.03,3.25) {Evaluator};
\coordinate (zoomT) at (ds.north east); \coordinate (zoomB) at (ds.south east);
\draw[inflow,teal] (4.35,4.85) to[out=180,in=90] (3.35,3.80);

\node[thinbox=sandEdge,minimum width=1.5cm,minimum height=0.95cm] (env) at (1.80,1.525) {};
\pic[scale=0.9] at (1.80,1.65) {db=sand};
\node[lane] at (1.80,1.20) {Environment};

\draw[callf] (ag.south -| 2.18,0) -- (2.18,3.37) -- (ds.west |- 0,3.37);
\node[lab,text=agentC,rotate=90,anchor=south] at (2.06,4.31) {tool call $\alpha_i$};
\draw[fwd] (ds.south) -- (3.03,2.35) -- (2.45,2.35) -- (env.north -| 2.45,0);
\node[lab,text=okC,anchor=west,inner sep=0.5pt] at (2.53,2.18) {forward};
\node[card=slateEdge,btitle,minimum height=0.5cm,anchor=center,inner sep=2.5pt] (lb) at (1.30,3.25)
  {\tikz[baseline=-0.6ex]\pic[scale=0.5]{robot=slate};\hspace{0.08cm}Labeler};
\draw[outp] (env.north -| 1.30,0) -- (lb.south);
\draw[outp] (lb.north) -- (ag.south -| 1.30,0);
\node[lab,text=sand,rotate=90,anchor=south] at (1.18,4.31) {output $o_i$};
\draw[flow] (lb.east |- 0,3.13) -- (ds.west |- 0,3.13);
\draw[flow,draw=inputC] (0.10,6.30) -- (0.10,3.25) -- (lb.west);
\draw[advice] (ds.north) -- (3.03,5.30) -- (ag.east |- 0,5.30);
\node[lab,text=hintC,anchor=south west] at ($(ag.east)+(0,0.1)$) {advice};

\begin{scope}[xshift=0.75cm]
\begin{scope}[on background layer]
  \path[shade,shading=axis,top color=tealTop,bottom color=tealBot,rounded corners=6pt] (4.15,0.95) rectangle (13.45,4.40);
  \draw[tealEdge,line width=0.6pt,rounded corners=6pt] (4.15,0.95) rectangle (13.45,4.40);
\end{scope}
\draw[tealEdge,line width=0.4pt] (zoomT) -- (4.15,4.40);
\draw[tealEdge,line width=0.4pt] (zoomB) -- (4.15,0.95);
\node[anchor=north west,text=teal,inner sep=2pt,text width=3cm] at (4.33,4.40) {\textbf{Tool call evaluator}\\(deterministic)};
\node[card=agentEdge,minimum width=1.4cm,minimum height=1.15cm,anchor=west,align=center,inner sep=3pt,font=\ttfamily\fontsize{5.6}{6.6}\selectfont] (ex) at (4.35,2.70)
  {\textcolor{muted}{\sffamily\fontsize{5.2}{6}\selectfont tool call $\alpha_i$}\\[1pt]
   \textcolor{ltlC}{book\_flight}\\[-1pt]
   \textcolor{smtC}{(class=}\\[-1pt]
   \textcolor{smtC}{business)}};
\draw[callf] ($(ex.west)+(-0.45,0.20)$) -- (ex.west |- 0,2.90);  %
\draw[flow,draw=ltlC] (ex.east |- 0,2.90) -- (6.05,2.90) |- (6.40,3.675);
\node[lab,anchor=north east,text=ltlC] at (5.99,3.62) {tool $t$};
\draw[flow,draw=smtC] (ex.east |- 0,2.50) -- (6.05,2.50) |- (6.40,1.80);
\node[lab,anchor=north east,text=smtC] at (5.99,1.78) {arguments $a$};
\node[chkcard=ltlEdge] (tpl) at (7.35,3.675)
  {\textcolor{ltlC}{Template (LTL)}\\[1pt]
   tool from a\\\textbf{finite} set\\[1pt]
   safe?\quad progress?};
\node[chkcard=smtEdge] (slv) at (7.35,1.80)
  {\textcolor{smtC}{Solver (SMT)}\\[1pt]
   arguments from\\an \textbf{infinite} domain\\[1pt]
   arguments ok?};
\node[circle,draw=teal,fill=white,line width=0.6pt,inner sep=0.8pt,font=\fontsize{7}{8}\selectfont,text=teal] (m) at (8.50,2.70) {$\wedge$};
\draw[flow,draw=ltlC] (tpl.east) -| (m.north);
\draw[flow,draw=smtC] (slv.east) -| (m.south);
\node[kite] (k1) at (10.35,3.95) {\textcolor{ltlC}{safe, progress}\\\textcolor{smtC}{arguments ok}};
\node[kite] (k2) at (10.35,2.70) {\textcolor{ltlC}{safe, no progress}\\\textcolor{smtC}{arguments ok}};
\node[kite] (k3) at (10.35,1.45) {\textcolor{ltlC}{unsafe} or\\\textcolor{smtC}{arguments wrong}};
\draw[flow,draw=teal] (m.east) -- (9.00,2.70) |- (k1.west);
\draw[flow,draw=teal] (m.east) -- (k2.west);
\draw[flow,draw=teal] (m.east) -- (9.00,2.70) |- (k3.west);
\node[okpill]    (v1) at (12.40,3.95) {\marktick\hspace{0.08cm}Allowed};
\node[okpill]    (v2) at (12.40,2.70) {\marktick\hspace{0.08cm}Allowed};
\node[blockpill] (v3) at (12.40,1.45) {\markcross\hspace{0.08cm}Blocked};
\draw[flow,draw=teal] (k1.east) -- (v1.west);
\draw[flow,draw=teal] (k2.east) -- (v2.west);
\draw[flow,draw=teal] (k3.east) -- (v3.west);
\draw[fwd] (v1.east) -- (13.65,3.95);
\draw[fwd] (v2.east) -- (13.65,2.70);
\draw[advthin] (v2.north) -- (12.40,3.15) -- (13.65,3.15);
\node[lab,anchor=south west,align=left,text=hintC] at (11.55,3.17) {advice:\\progressing call};
\draw[blockC,line width=0.9pt] (v3.east) -- (13.21,1.45);
\draw[blockC,line width=1.1pt,line cap=round] (13.22,1.37) -- (13.38,1.53) (13.22,1.53) -- (13.38,1.37);
\draw[advthin] (v3.north) -- (12.40,1.90) -- (13.65,1.90);
\node[lab,anchor=south west,align=left,text=hintC] at (11.55,1.92) {advice: enabling call\\or repaired argument};
\end{scope}
\end{tikzpicture}
\caption{Overview of \framework. Manual and tool schema are formalized into an LTL-based template and an SMT-based solver (offline, top right), used to shield every tool call (online, left) by deterministically evaluating (bottom right) if tool calls are safe and progressing (violet, LTL), and use correct arguments (blue, SMT); unsafe and non-progressing calls receive advice (orange).}
\label{fig:pipeline}
\end{figure*}
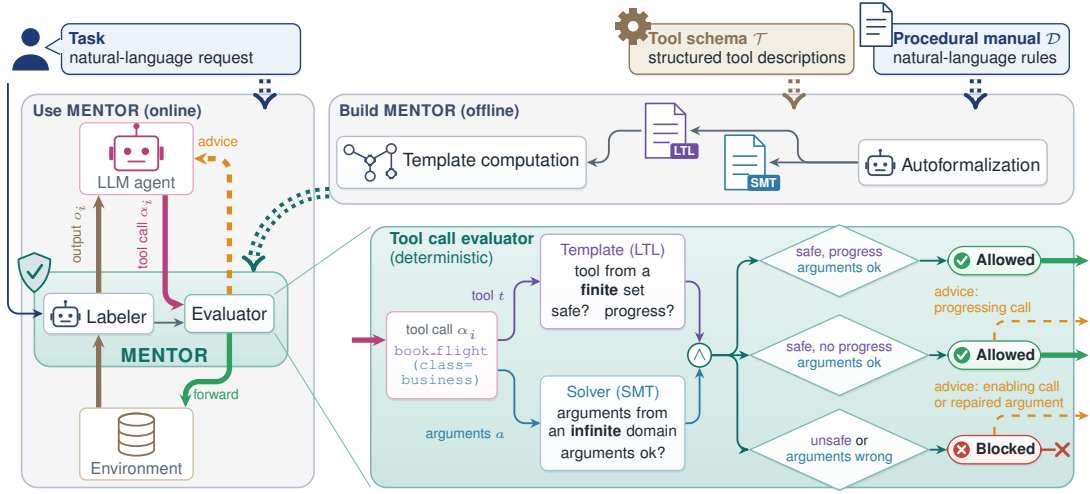

LTL was recently proposed as the right formalism to capture procedural dependencies~\citep{anand2026mantra,aminof2026specifying}, and translating natural language into LTL is well studied~\citep{chen2023nl2tl,cosler2023nl2spec,fuggitti2023nl2ltl,liu2023grounding,ma2025bridging,wang2025conformalnl2ltl}.
However, none of the above work used LTL in the context of shielding LLM agents. 
In addition, LTL constraints are propositional and cannot express numeric ranges or unbounded argument domains (\chal{2}). Auto-generated SMT-based constraints have so far only served post-hoc verification~\citep{an2025neurosymbolic,anand2026mantra}, not runtime shielding.

\section{Overview}\label{sec:overview}

\framework (\emph{Monitoring and ENforcement of Temporal Objectives at Runtime}) sits between a tool-using LLM agent and its environment (Fig.~\ref{fig:pipeline}). We outline how its components address \chal{1}--\chal{4}.

\paragraph{Setting.}
A tool schema $\tool$ defines the set of tools available to the agent.
Let $\mathcal{A}$ be the set of valid tool-calls $\alpha=(t,a)$, with tool $t\in\tool$ and argument assignment $a$, including the terminating call \code{done} that ends the interaction. Let $\mathcal{H}$ denote the set of finite interaction traces $h=[o_0,\alpha_1,o_1,\dots,\alpha_l,o_l]$, where $o_0$ is the natural-language task and $o_i$ the environment's response to $\alpha_i$. The agent $\pi:\mathcal{H}\to\Delta(\mathcal{A})$ repeatedly samples $\alpha_i\sim\pi(\cdot\mid h)$ and observes $o_i$, until it generates \code{done}.
Each call is generated internally by an autoregressive LLM over a token vocabulary $V$.
A natural-language procedural manual $\doc$ specifies which traces are acceptable.

\paragraph{Deterministic tool call evaluator (Fig.~\ref{fig:pipeline}, bottom).}
Our goal is that the trace generated by the agent completes the task $o_0$ while complying with $\doc$. 
To this end, for each proposed call $\alpha_i$ and history $h$, the evaluator asks two questions: is $\alpha_i$ \emph{safe}, i.e., whether a compliant complete trace under $\doc$ still exists after $\alpha_i$, and does $\alpha_i$ make \emph{progress} towards completion (\chal{1})? Also, are its \emph{arguments} $a_i$ permitted (\chal{2})? 
Safe calls with permitted arguments are executed; non-progressing calls additionally receive advice identifying a progressing call. Other calls are blocked, returning advice identifying a prerequisite call or repaired arguments instead. The call \code{done} is blocked while the task remains incomplete (Fig.~\ref{fig:pipeline}, bottom).

\paragraph{Building \framework offline (Fig.~\ref{fig:pipeline}, top).}
\framework is constructed once per manual $\doc$ and tool schema $\tool$ and then reused for every task. First, a multi-agent \emph{autoformalization} loop (Sec.~\ref{sec:autoformalization}) translates $\doc$ into a formal specification with two parts (Sec.~\ref{sec:spec}): (i) \emph{temporal requirements} in linear temporal logic (LTL) formulas to capture tool call order, restrictions that tool outputs impose, and obligations before \code{done},
and (ii) \emph{argument constraints} in satisfiability modulo theories (SMT) encoding, which fix the permitted argument values over possibly unbounded argument domains (\chal{2}).
The LTL requirements are compiled into a game between agent and environment on a finite graph, and a \emph{strategy template} (Sec.~\ref{sec:shield-construction}) marks, at every vertex of this graph, which calls are unsafe and which are progressive
(\chal{1}).
The template and SMT solver form the deterministic evaluator.

\paragraph{Applying \framework online (Fig.~\ref{fig:pipeline}, left).}
Since the task $o_0$ and some tool outputs are in natural language, at runtime, a \emph{labeler} LLM (Sec.~\ref{sec:online}) grounds them into formal, \emph{labeled events} consistent with the specifications (\chal{3}).
For each proposed call, the shield advances through the game graph following $h$, consults the template and the solver, and produces a verdict and advice.
These reach the agent in one of two ways (\chal{4}):
(i) via a \emph{prompt shield} that delivers them as text in the tool output,
or (ii) via a \emph{token shield} that additionally masks the tokens of unsafe calls and boosts those of progressing calls during generation.

\section{Methodology}\label{sec:methodology}
This section formalizes the components in Fig.~\ref{fig:pipeline}
and explains how each addresses challenges \chal{1}--\chal{4}.
We introduce our specification language in Sec.~\ref{sec:spec},
then describe the offline construction of \framework in
Sec.~\ref{sec:autoformalization} and its online use for
procedural shielding in Sec.~\ref{sec:online}.

\subsection{Specification Language}\label{sec:spec}

\paragraph{Temporal Requirements via LTL.}
LTL extends propositional logic with temporal operators that express requirements on the ordering of events over time. We use standard LTL syntax with finte trace semantics~\citep{de2013linear} (refer to App.~\ref{app:spec-language} for details).
We interpret these requirements over interaction traces $h$. At each step, propositions are evaluated from predicates on the tool call $\alpha=(t,a)$ and, when needed, its observed outcome $o$. Formulas can thus relate proposed actions to
preceding calls and results, and require successful completion before termination.

In our example in Fig.~\ref{fig:shield-example}, booking travel requires a preceding distance check, the resulting distance restricts the permitted travel mode, and the requested booking must eventually succeed. 
For origin $x$ and destination $y$, let
\ltlform{\ltlprop{checked\_distance}(x,y)} denote a \code{check\_distance} tool call, and \ltlform{\ltlprop{booked\_train}(x,y)} and \ltlform{\ltlprop{booked\_flight}(x,y)} the corresponding booking calls. The distance-check prerequisite P1 can be expressed in LTL as
\[
\ltlform{
  \ltlnot\bigl(
    \ltlprop{booked\_train}(x,y)
    \ltlor \ltlprop{booked\_flight}(x,y)
  \bigr)
  \ltlW
  \ltlprop{checked\_distance}(x,y)
}.
\]
Here, $\ltlW$ denotes `weak until', which means that bookings are forbidden until the distance check occurs, and throughout the interaction if it never occurs. This blocks the bookings in steps~1 and~2. Crucially, each $(x,y)$ is tracked separately, i.e., a Toronto--Chicago check cannot satisfy the prerequisite for Toronto--Montreal.

Tool results can further restrict the continuation. Let \ltlform{\ltlprop{distance\_small}(x,y)} hold if \code{check\_distance} returns a distance below 500\,km for that route. The prohibition on flights below 500\,km in P2 can then be expressed in LTL as 
$\ltlform{
  \ltlG\bigl(
    \ltlprop{distance\_small}(x,y)
    \ltlimplies
    \ltlG\,\ltlnot\ltlprop{booked\_flight}(x,y)
  \bigr)}$. Here, $\ltlG$ denotes `globally', i.e., a result below 500\,km forbids \emph{all} subsequent flight bookings for that route.

Both formulas above would permit an execution that checks the distance and then stops without making a booking,
so we also express \emph{obligations} to complete the task.
Let \ltlform{\ltlprop{booked\_train\_ok}(x,y)} and \ltlform{\ltlprop{booked\_flight\_ok}(x,y)} denote booking calls whose outcomes confirm success. For the requested route, the requirement is 
$\ltlform{
  \ltlF\bigl(
    \ltlprop{booked\_train\_ok}(x,y)
    \ltlor \ltlprop{booked\_flight\_ok}(x,y)
  \bigr)
}$, where $\ltlF$ denotes `eventually' and requires a successful booking before termination. An activated obligation that remains unfulfilled is called an \emph{open duty}. Here, the requested booking is an open duty from the start, remains open after the distance check in step~3 and gets fulfilled by the confirmed booking in step~5.

These examples illustrate how  our temporal specification captures both \emph{safety}, by excluding forbidden behavior, and \emph{progress}, by requiring obligations to be fulfilled before termination.
Grounding propositions in tool calls and their outcomes enables runtime monitoring.
However, tool arguments may range over unbounded domains, so we parameterize the propositions arguments and bind them to concrete entities at runtime.
Preserving these bindings is essential, as ignoring $(x,y)$ would make a short-distance result for one route prohibit flights on every route. To mitigate this, we use SMT to check that the tool calls' arguments are consistent with the proposition bindings.

\paragraph{Argument Constraints via SMT.}
We express \emph{argument constraints} as predicates on the arguments $a$ of a call $\alpha=(t,a)$, which may also depend on information from the interaction trace $h$. Each constraint applies to calls identified by its associated proposition. In our example, satisfying the distance-check does not by itself permit a business-class booking, since P3 also requires the agent to establish an urgent reason and the absence of an economy seat for that route.

We encode argument constraints using SMT, which decides the satisfiability of first-order formulas modulo theories. In our implementation, we use Z3~\citep{demoura2008z3} with quantifier-free formulas over strings and linear integer arithmetic.
SMT can directly express argument restrictions, enabling a solver to check whether multiple restrictions are jointly satisfiable and efficiently find satisfying values without fully enumerating the argument domain.
Our argument constraints use a domain-specific language that is evaluated directly on proposed arguments and has an SMT encoding for satisfiability checks and symbolic argument search (App.~\ref{app:spec-language}).

For the Toronto--Chicago flight in Fig.~\ref{fig:shield-example}, let $\code{business\_allowed}(h,x,y)$ indicate whether $h$ establishes both an urgent reason and the absence of an economy seat for route $(x,y)$. P3 constrains calls
satisfying \ltlform{\ltlprop{booked\_flight}(x,y)} via the constraint
\vspace{-1mm}
\begin{equation*}
  \vspace{-1mm}
\label{eq:travel-class}
  a[\code{class}]=\code{business}
  \ltlimplies
  \code{business\_allowed}(h,x,y),
\end{equation*}
where $a[\code{class}]$ denotes the booking-class argument represented as a string-valued SMT variable, and $\code{business\_allowed}(h,x,y)$ is a fixed Boolean value for the given history and route. The distance check in step~3 satisfies P1, and its 700\,km result does not trigger P2's flight prohibition. However, no urgent reason has been established, so $\code{business\_allowed}(h,\code{Toronto},\code{Chicago})$ is false. The business-class booking in step~4 therefore violates the constraint.

\subsection{Building \framework Offline}\label{sec:autoformalization}

To build \framework, we first translate natural-language manuals into the specifications of Sec.~\ref{sec:spec} (\chal{3}), to compute templates that provide the guidance required by \chal{1} (Fig.~\ref{fig:pipeline}, top right).

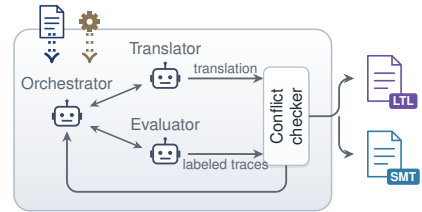
\begin{wrapfigure}{r}{0.4\textwidth}
\vspace{-\baselineskip}
\centering
\begin{tikzpicture}[x=1cm,y=1cm,font=\sffamily\fontsize{6.2}{7.2}\selectfont,
  /utils/exec={%
    \definecolor{ink}{HTML}{1F2933}%
    \definecolor{muted}{HTML}{5B6773}%
    \definecolor{slate}{HTML}{3B4C6B}%
    \definecolor{slateEdge}{HTML}{B4BFCF}%
    \definecolor{inputC}{HTML}{1F3B73}%
    \definecolor{inputFill}{HTML}{EEF2F9}%
    \definecolor{sand}{HTML}{8A7455}%
    \definecolor{sandFill}{HTML}{F3EEE4}%
    \definecolor{ltlC}{HTML}{6B4FA0}%
    \definecolor{smtC}{HTML}{2E7BA6}%
  },
  card/.style={draw=#1,shade,shading=axis,top color=white,bottom color=#1!6!white,
    line width=0.5pt,rounded corners=2.5pt,inner sep=3pt,text=ink,align=center,
    drop shadow={opacity=0.15,shadow xshift=0.4pt,shadow yshift=-0.9pt}},
  input/.style={draw=inputC,fill=inputFill,line width=0.5pt,rounded corners=3pt,inner sep=3pt,text=ink,align=left,
    drop shadow={opacity=0.15,shadow xshift=0.4pt,shadow yshift=-0.9pt}},
  panel/.style={draw=#1,shade,shading=axis,top color=white,bottom color=#1!35!white,rounded corners=6pt,line width=0.6pt},
  ptitle/.style={font=\sffamily\fontsize{7.2}{8.5}\selectfont\bfseries,inner sep=5pt},
  lab/.style={text=muted,font=\sffamily\fontsize{5.2}{6}\selectfont,inner sep=1.5pt},
  lane/.style={text=muted,font=\sffamily\fontsize{6.2}{7.2}\selectfont,align=center,inner sep=1pt},
  tag/.style={fill=#1,text=white,font=\sffamily\fontsize{4.6}{5}\selectfont\bfseries,inner sep=1.3pt,rounded corners=1pt},
  flow/.style={-{Stealth[length=4pt,width=3.5pt]},line width=0.7pt,draw=muted,rounded corners=4pt},
  inflow/.style={double,double distance=0.8pt,dotted,line width=1pt,-{Implies},draw=inputC},
  both/.style={{Stealth[length=3.5pt,width=3pt]}-{Stealth[length=3.5pt,width=3pt]},line width=0.6pt,draw=muted},
  pics/robot/.style={code={%
    \draw[fill=white,draw=#1,line width=0.7pt,rounded corners=1.5pt] (-0.32,-0.25) rectangle (0.32,0.25);
    \fill[#1] (-0.13,0.03) circle (0.055) (0.13,0.03) circle (0.055);
    \draw[#1,line width=0.7pt,line cap=round] (-0.12,-0.12) -- (0.12,-0.12);
    \draw[#1,line width=0.7pt] (0,0.25) -- (0,0.38); \fill[#1] (0,0.40) circle (0.045);
    \fill[#1,rounded corners=0.5pt] (-0.40,-0.10) rectangle (-0.34,0.10) (0.34,-0.10) rectangle (0.40,0.10);}},
  pics/gear/.style={code={%
    \foreach \a in {0,45,...,315}{\fill[#1,rotate=\a] (-0.07,0.22) rectangle (0.07,0.34);}
    \fill[#1] (0,0) circle (0.26); \fill[white] (0,0) circle (0.10);}},
  pics/doc/.style={code={%
    \path[fill=white,draw=#1,line width=0.6pt,line join=round]
      (-0.22,-0.30) -- (-0.22,0.30) -- (0.10,0.30) -- (0.22,0.18) -- (0.22,-0.30) -- cycle;
    \draw[#1,line width=0.6pt] (0.10,0.30) -- (0.10,0.18) -- (0.22,0.18);
    \draw[#1,line width=0.6pt,line cap=round] (-0.13,0.07) -- (0.13,0.07) (-0.13,-0.03) -- (0.13,-0.03) (-0.13,-0.13) -- (0.04,-0.13);}},
]

\draw[panel=slateEdge] (0.00,0.55) rectangle (4.20,2.95);
\pic[scale=0.7] at (0.50,3.05) {doc=inputC};
\pic[scale=0.4] at (1.0,3.05) {gear=sand};
\draw[inflow] (0.50,2.9) -- (0.50,2.50);
\draw[inflow,draw=sand] (1.0,2.9) -- (1.0,2.50);
\pic[scale=0.5] at (0.70,1.80) {robot=slate};  \node[lane] at (0.70,2.24) {Orchestrator};
\pic[scale=0.5] at (2.00,2.30) {robot=slate};  \node[lane] at (2.00,2.70) {Translator};
\pic[scale=0.5] at (2.00,1.30) {robot=slate};  \node[lane] at (2.00,1.70) {Evaluator};
\node[card=slateEdge,rotate=90,minimum width=1.3cm,minimum height=0.6cm,inner sep=2pt,align=center] (cc) at (3.60,1.80) {Conflict\\checker};
\draw[both] (1.00,1.93) -- (1.70,2.22);
\draw[both] (1.00,1.67) -- (1.70,1.38);
\draw[flow] (2.30,2.30) -- node[lab,above] {translation} (3.30,2.30);
\draw[flow] (2.30,1.30) -- node[lab,below] {labeled traces} (3.30,1.30);
\draw[flow] (3.60,1.15) -- (3.60,0.80) -- (0.70,0.80) -- (0.70,1.60);

\draw[flow] (3.90,1.80) -- (4.30,1.80) |- (4.50,2.30);
\draw[flow] (4.30,1.80) |- (4.50,1.30);
\pic[scale=1.0] at (4.90,2.30) {doc=ltlC};  \node[tag=ltlC] at (5.15,2.00) {LTL};
\pic[scale=1.0] at (4.90,1.30) {doc=smtC};  \node[tag=smtC] at (5.15,1.00) {SMT};
\end{tikzpicture}
\vspace{-0.5\baselineskip}
\caption{Autoformalization.}
\label{fig:autoformalization}
\vspace{-\baselineskip}
\end{wrapfigure}
\paragraph{Autoformalization.}
We use a multi-agent refinement loop with three LLM agents: a translator, an evaluator, and an orchestrator (Fig.~\ref{fig:autoformalization}). The \emph{translator} takes the manual and the tool schema and formalizes the requirements in the specification language defined above. The \emph{evaluator} generates test traces from the manual, consisting of tool calls $\alpha_i$ and their outcomes $o_i$, and labels them as compliant or non-compliant with the manual.
A deterministic conflict checker then compares the proposed specification's verdicts on these traces with the evaluator's proposed labels and returns disagreements to first the translator to revise the specification.
The translator can contest a label, by providing an explanation grounded in the manual, which the evaluator reviews.
The \emph{orchestrator} coordinates these revisions and resolves remaining disagreements referring to the manual. The loop ends when the specification agrees with the evaluated traces or reaches a fixed iteration limit.

Note that if the loop terminates successfully, it provides some evidence of consistency between the generated specification and the manual, but it does not prove that it captures every requirement correctly.
However, formalizing natural-language requirements is difficult even for human experts~\citep{greenman2023little,czepa2019modeling}.
Given that there exists no perfect method for this task, we believe that our loop provides a practical way to iteratively refine the specification and will become increasingly reliable as LLMs and autoformalization techniques improve.

\begin{wrapfigure}{r}{0.45\textwidth}
\vspace{-\baselineskip}
\centering
\begin{tikzpicture}[x=1cm,y=1cm,font=\sffamily\fontsize{6.2}{7.2}\selectfont,
  /utils/exec={%
    \definecolor{ink}{HTML}{2E2E2E}%
    \definecolor{muted}{HTML}{6B6B6B}%
    \definecolor{agentC}{HTML}{B8407A}%
    \definecolor{agentFill}{HTML}{FBEEF4}%
    \definecolor{sand}{HTML}{8A7455}%
    \definecolor{toolFill}{HTML}{F3EEE4}%
    \definecolor{blockEdge}{HTML}{C0392B}%
    \definecolor{blockFill}{HTML}{FDECEA}%
    \definecolor{hintEdge}{HTML}{D68910}%
    \definecolor{okEdge}{HTML}{1E8449}%
    \definecolor{okFill}{HTML}{E8F6EC}%
    \def\stepmark#1{\textcolor{muted}{\ding{#1}}}%
  },
  sys/.style={circle,draw=agentC,fill=agentFill,line width=0.6pt,minimum size=0.42cm,inner sep=0pt,text=ink},
  env/.style={rectangle,draw=sand,fill=toolFill,line width=0.6pt,minimum size=0.38cm,inner sep=0pt},
  bad/.style={rectangle,rounded corners=1.5pt,draw=blockEdge,fill=blockFill,line width=0.6pt,minimum size=0.36cm,inner sep=1pt,text=blockEdge},
  good/.style={circle,draw=okEdge,fill=okFill,line width=0.6pt,minimum size=0.42cm,inner sep=0pt,text=okEdge},
  acc/.style={circle,double,double distance=1pt,draw=okEdge,fill=okFill,line width=0.5pt,minimum size=0.40cm,inner sep=0pt,text=okEdge},
  res/.style={text=black,font=\sffamily\fontsize{5.4}{6.0}\selectfont,align=center,inner sep=1.5pt},
  lab/.style={font=\ttfamily\fontsize{5.4}{6.0}\selectfont,inner sep=1.2pt,text=ink,align=center},
  unsafe/.style={-{Stealth[length=3pt]},blockEdge,dashed,line width=0.7pt},
  prog/.style={-{Stealth[length=3pt]},hintEdge,line width=0.9pt},
  safe/.style={-{Stealth[length=3pt]},muted,line width=0.5pt},
]
\node[sys]  (v0)  at (0.35,2.30) {$v_0$};
\node[bad]  (bot) at (0.35,3.55) {$\bot$};
\node[bad]  (bot2) at (0.35,1.05) {$\bot$};
\node[env]  (e0)  at (2.45,2.30) {};
\node[sys]  (v1)  at (2.90,3.55) {$v_1$};
\node[sys]  (v2)  at (2.90,1.05) {$v_2$};
\node[env]  (e1)  at (4.90,3.55) {};
\node[env]  (e2)  at (4.90,1.05) {};
\node[sys]  (v3)  at (5.40,2.30) {$v_3$};
\node[acc]  (acc) at (3.9,2.3) {\ding{51}};
\node[res,anchor=north] at (v0.south)  {P1 open\\booking due};
\node[res,anchor=south] at (v1.north)  {train only\\booking due};
\node[res,anchor=north] at (v2.south) {booking due};
\node[res,anchor=south] at (bot.north) {violation};
\node[res,anchor=north] at (bot2.south) {violation};
\draw[unsafe] (v0) -- node[lab,right,yshift=4pt] {book\_train\,\stepmark{182}\\book\_flight\,\stepmark{183}\\done} (bot);
\draw[prog]   (v0) -- node[lab,above,yshift=2.5pt] {check\_distance\,\stepmark{184}} (e0);
\draw[safe]   (e0) -- node[lab,above,sloped,pos=0.5] {<500\,km} (v1);
\draw[safe]   (e0) -- node[lab,below,sloped,pos=0.5] {$\geq$500\,km} (v2);
\draw[unsafe] (v1) -- node[lab,above,sloped] {book\_flight\\ done} (bot);
\draw[unsafe] (v2) -- node[lab,below,pos=0.7] {done} (bot2);
\draw[prog]   (v1) -- node[lab,anchor=north,pos=0.5,yshift=-3pt] {book\_train} (e1);
\draw[prog]   (v2) -- node[lab,anchor=south east,pos=1,xshift=2pt,yshift=3pt] {book\_flight\,\stepmark{185}\stepmark{186}\\book\_train} (e2);
\draw[safe]   (v2) to[out=160,in=230,looseness=1] node[lab,below left,pos=0.5,xshift=0pt,yshift=5pt] {check\_\\distance} (e0);
\draw[safe]   (e1) -- node[lab,below,sloped] {CONF.} (v3);
\draw[safe]   (e2) -- node[lab,above,sloped] {CONF.} (v3);
\draw[prog]   (v3) -- node[lab,above] {done} (acc);
\draw[safe]   (e1) to[bend right] node[lab,above,sloped] {DENY} (v1);
\draw[safe]   (e2) to[bend left] node[lab,below,sloped] {DENY} (v2);
\end{tikzpicture}
\vspace{-0.5\baselineskip}
\vspace{-1mm}
\caption{Game graph and template (excerpt) for the example from Fig.~\ref{fig:shield-example}.}
\label{fig:template-example}
\vspace{-0.5\baselineskip}
\end{wrapfigure}
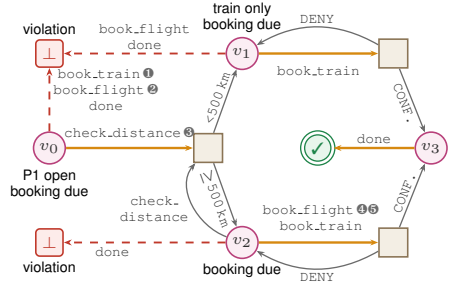

\paragraph{Template Construction.}\label{sec:shield-construction}
The LTL specification $\varphi$ defines compliance for interaction traces $h$.
Shielding, however, requires determining which tool calls must be blocked or advised \emph{during} interaction, before the trace is complete.
Towards this, we adopt a standard game-theoretic approach, which automatically maps $\varphi$ into a finite two-player game between the agent and the environment (see Fig.~\ref{fig:template-example} for an example)~\citep{PnueliRosner1989}. 
At an agent vertex (circle), the agent chooses a call, which leads to an environment vertex (square), at which the environment chooses its outcome. Each vertex abstracts the part of the history $h$ relevant for the specification into residual LTL formulas, i.e., the requirements that the remaining interaction must still satisfy (black).
Calls violating a requirement lead to a sink $\bot$ (red), from which no compliant continuation exists, and \code{done} leads to the accepting vertex (\ding{51}, green) if and only if no duty is open.

To obtain a shielding mechanism from the above translation, we employ a recent method
~\citep{anand2023synthesizing,anand2026follow} that automatically computes a \emph{strategy template} over a given game. At each agent vertex, the template marks (i) \emph{unsafe} tools (red, dashed), which violate a requirement immediately or later unavoidably, and (ii) \emph{progressing} tools (orange, solid), which advance the fulfillment of open duties. In Fig.~\ref{fig:template-example}, \code{done} is unsafe at $v_0$, $v_1$ and $v_2$ because the booking duty is still open, but it is progressing at $v_3$ because the confirmed booking fulfills it.

Intuitively, the template turns the global requirements into local constraints that let \framework block unsafe calls and advise progress toward compliance.
It is also permissive, because the agent may choose freely any safe tool, including non-progressing calls.
\framework interferes with the agent's preferences only where compliance requires it, and reminds of the open duties with its advice so that the agent can choose when to address them.
However, the template only ranges over tools.
The admissibility of their \emph{arguments} requires additional SMT-based checks. In Fig.~\ref{fig:template-example}, booking a business-class flight (step~4 from Fig.~\ref{fig:shield-example}) is progressing for the template in $v_2$ but blocked by the argument constraint P3. The next section discusses how this is handled by \framework at runtime.

\subsection{Applying \framework at Runtime}\label{sec:online}

The template-annotated game (Fig.~\ref{fig:template-example}) provides the \emph{context-dependent} behavior of \framework.
At runtime, the executed tool calls and outcomes are followed through the game, and the template of the current vertex determines unsafe and progressing calls.
\framework additionally checks the \emph{arguments} of proposed calls against the SMT constraints and, when a call is blocked, uses the solver to compute admissible values from the history $h$.
Together, this yields the evaluator of Fig.~\ref{fig:pipeline} (bottom).

Two further components make this deterministic evaluator applicable to real agent interactions: (i) a labeler LLM that grounds propositions in the natural-language task description and tool outputs (the runtime part of \chal{3}), and (ii) two shielding mechanisms, a prompt shield and a token shield, that deliver the evaluator's verdicts and advice to the agent (\chal{4}). We describe both below.

\paragraph{Labeler LLM.}

The tool-call evaluator operates on structured data, applying predicates to arguments $a$ and outcomes $o$ in $h$ to evaluate propositions such as \ltlform{\ltlprop{booked\_flight}(x,y)} and argument constraints such as P3. However, the task description $o_0$ and free-text tool outputs, e.g.\ from \code{check\_distance}, are natural language. \framework uses a  \emph{labeler LLM} to convert both into structured events according to the specification, on which the tool-call evaluator deterministically makes shielding decisions.

Before the first tool call, the labeler turns $o_0$ into one or more \emph{pseudo tool calls}. The specification declares their names, arguments, and a plain-text extraction rule per argument. For Fig.~\ref{fig:shield-example}, the task becomes \code{request\_travel(Toronto, Chicago, class=business, urgent=false)}.
The predicates of the temporal requirements bind $x=\code{Toronto}$, $y=\code{Chicago}$ to this event, which activates the route's booking duty, and no urgent reason is given which makes $\code{business\_allowed}(h,x,y)$ false.

Further, for each tool output marked as natural language, the specification declares a small label vocabulary. After a tool call, the labeler maps the returned text to exactly one label or none, and stores it in $o_i$ for propositions to read like any other value. In Fig.~\ref{fig:shield-example}, \code{check\_distance}
is labeled as  $\geq$500\,km,
determining that \ltlform{\ltlprop{distance\_small}(x,y)} does not hold and the agent moves to $v_2$ in
Fig.~\ref{fig:template-example}.

\paragraph{Prompt shield.}%
The tool-call evaluator enforces its decision through the tool call output, and the prompt shield delivers the verdict and advice to the agent in the form of text along with the tool outcome.
If a call $\alpha=(t,a)$ is blocked, it is not executed, and the agent receives as feedback the violated excerpt from the manual and a suggested tool call.
The suggestion may be either
the template-indicated missing prerequisite call at the current vertex,
or an SMT-repaired call
for a violated argument constraint.
Alternatively, if $\alpha$ is allowed, it is executed.
However, if duties remain open, its result is augmented with a reminder listing the open duties and a progressing call.
The call \code{done} is refused while duties remain open, or until a preset limit on the number of calls is reached.

As an example, consider that at $v_0$ of Fig.~\ref{fig:template-example} the agent calls \code{book\_train} (step~1 of Fig.~\ref{fig:shield-example}).
This call is unsafe at $v_0$ as \code{check\_distance} is a prerequisite, so the agent receives the feedback: 
\vspace{-1mm}
\begin{Verbatim}[fontsize=\scriptsize,breaklines,breakanywhere,xleftmargin=1em]
[SHIELD] Blocked book_train(origin="Toronto", destination="Chicago").
- Policy rule for trip Toronto->Chicago: Check the distance before any booking. (P1)
  First: check_distance(origin="Toronto", destination="Chicago")
\end{Verbatim}
\vspace{-1mm}

The prompt shield relies on agents following its feedback, which may be difficult for small open-weight models as shown in our experiments in Sec.~\ref{sec:experiments}.
If the next-token distribution is accessible, \framework can additionally apply a token shield to enforce the same constraints at the token level.

\paragraph{Token shield.}%
LLM agents express tool calls as token sequences, so we modify the next-token distribution during generation of the tool name $t$ and arguments $a$.
We do this by adapting the technique from
\textsc{STARS}~\citep{anand2026follow}, which uses a strategy template to modify an RL policy's action distribution, assigning unsafe actions probability zero and increasing
the probability of progressing actions.

Before each model call, \framework derives unsafe and progressing tool names and argument values from the current tool-call evaluation.
Since tool names precede arguments, a name is considered unsafe only if every call with that tool name is unsafe. At $v_0$ in Fig.~\ref{fig:template-example}, \code{book\_train}, \code{book\_flight}, and \code{done}
are unsafe, and \code{check\_distance} is progressing.
For arguments, SMT identifies unsafe candidate values from the history $h$ and suggested calls from prior advice. At $v_2$, \code{business} from the task description is unsafe under P3, while \code{economy} from the advised call is progressing.
Each name and value is tokenized according to the model's tool-call format to identify its token sequence, so that we do not change the token distribution outside of tool calls.

Let $V$ be the token vocabulary, $\boldsymbol{\tau}_{\mathrm{full}}$ the token context, and $p(\tau\mid\boldsymbol{\tau}_{\mathrm{full}})$ the next-token probability for $\tau\in V$. 
Since tool names may span several tokens and share prefixes, e.g., \code{book\_flight} and \code{book\_train},
we mask a token if the tokens written so far followed by $\tau$ can continue into unsafe names alone and into no safe one.
For argument values, we only mask tokens completing an identified unsafe value, to avoid overrestriction. Tokens that continue a progressing call are
\emph{boosted}, including prefixes shared with safe, non-progressing alternatives.
The modified distribution is
\vspace{-1mm}
\begin{equation*}
  \vspace{-1mm}
\widetilde{p}(\tau\mid\boldsymbol{\tau}_{\mathrm{full}})
=
\begin{cases}
0,
  & \text{if $\tau$ is masked},\\
p(\tau\mid\boldsymbol{\tau}_{\mathrm{full}})
  + \eta b_{\tau},
  & \text{if $\tau$ is boosted},\\
p(\tau\mid\boldsymbol{\tau}_{\mathrm{full}}),
  & \text{otherwise},
\end{cases}
\end{equation*}
renormalized before sampling. Here, $\eta>0$ controls the boost strength and $b_\tau$ a counter that starts at $1$ for each progressing tool call and increases for each step it is not called; if several progressing tools share $\tau$, the largest counter is used.
At $v_2$, the agent may still call \code{check\_distance}, but each such call further boosts \code{book\_flight}.

After the tool call has been generated, it is still evaluated through the deterministic tool call evaluator and its advice returned to the LLM agent, because the modified distribution does not guarantee compliance on its own.
That is because each argument is constrained separately during generation, so unsafe combinations of arguments may still be generated.

\section{Experiments}\label{sec:experiments}

We implemented \framework and evaluated its performance on three compliance benchmarks.
We used Claude Opus 5 for the autoformalization loop, and tried several open-weights and proprietary models as LLM agents and as labelers.
For the SMT solver, we used Z3~\citep{demoura2008z3}.
All experiments were run on a cluster with nodes of 2x Intel Xeon Gold 5317 CPUs, 2x Nvidia A100 GPUs and 1024\,GB RAM.
See App.~\ref{app:more-experiments} for the exact model versions and all hyperparameters used.

\paragraph{Benchmarks.}
We evaluated \framework on
\mantra~\citep{anand2026mantra}, SOP-Bench~\citep{nandi2026sop} and \textsc{AgentOrca}~\citep{li2025sopbench}.
Each benchmark contains several domains with a procedural manual, tool schema, tasks, and a compliance evaluator.
We evaluated \framework on 4 domains from \mantra (2 original and 2 adapted from \taubench), 2 domains from SOP-Bench, and 2 domains from \textsc{AgentOrca}.
However, we observed that SOP-Bench's evaluator does not read the arguments of tool calls and misses argument-based constraint violations.
For this reason, we did not include SOP-bench in our main results, and refer to App.~\ref{app:sopbench} for further details.

\paragraph{Baselines.}
We compare the compliance of agents shielded with both versions of \framework (prompt and token shield) against unshielded agents and, on \mantra, three other baselines.
\textsc{DynaGuard} provides a specialized LLM that judges the compliance of tool calls at runtime.
\textsc{Progent} provides a specialized LLM that dynamically creates constraints during the interaction, which are then deterministically checked against the tool calls.
\textsc{AgentSpec} provides a DSL in which humans write constraints that are then enforced deterministically at runtime, and a manually written labeler that uses fixed keywords and regex patterns.
We write the constraints and labeling rules for \textsc{AgentSpec} using \framework's autoformalization loop (see App.~\ref{app:ablations}), and in Table~\ref{tab:llmgen-passsupk} we show that it is comparable with \textsc{AgentSpec}'s own LLM-based rule generator that uses $10\%$ of the test cases as a validation set.

\paragraph{Metric.}
We ran $5$ independent runs of each task per agent and report \passSupFive{}, the probability that all $5$ runs of a task succeed, averaged over the entire benchmark.
In App.~\ref{app:more-experiments}, we also report \passAtOne{} and \passAtFive{}, the average single-run success rate and the probability that at least one of $5$ runs of a task succeeds, respectively.

\paragraph{Results.}
Table~\ref{tab:passsup5-by-domain} summarizes the results for both benchmarks.
\framework consistently improves compliance for every agent, proprietary and open-weights, in comparison to unshielded agents, and outperforms the baselines.
We observe that the prompt-based shield is most effective for stronger agents that can act on its feedback.
On \mantra, it sometimes struggles to help Llama~3.1 (see App.~\ref{app:prompt-vs-token} for an example), but the token-based shield instead improves it from $\sim 5 - 20\%$ to $\sim 30 - 60\%$ in all domains.
The baselines are less effective, often due to their lack of guidance towards progress, resulting in agents repeating blocked calls over and over (see App.~\ref{app:failure-traces-baselines} for examples).

\paragraph{Impact of labeler and autoformalization.}
In Table~\ref{tab:passsup5-by-domain} the autoformalization was done by Claude Opus 5, and the labeler was gpt-4.1 for the open-weights models and gpt-4.1-mini for the proprietary models.
To understand the role of these choices, we conducted ablation studies on \mantra in App.~\ref{app:ablations}, varying the autoformalization and labeling models and observing their impact on compliance.
The results in Table~\ref{tab:ablation-translator-mantra-passhatk}
reveal that strong models are necessary for autoformalization using our loop, otherwise compliance drops.
Similarly, the results in Table~\ref{tab:ablation-labeler-mantra-passhatk}
show that open-weights models struggle to label successfully, but even small proprietary models such as gpt-4.1-mini are satisfactory.
However, a small open-weights model can be easily finetuned to match this performance and act as a generalizable labeler, as we demonstrate with Qwen3-4B in Table~\ref{tab:shield-comparison-passsup5} in App.~\ref{app:finetune-labeler}, where we trained it on gpt-4.1 labels from one domain and tested it on others.

\paragraph{Cost and latency.} 
The autoformalization is performed once per manual and incurs a one-time cost, after which the generated constraints can be reused for all tasks within that domain. For this step, we used a 100 USD per month Claude subscription.
The most expensive labeler, gpt-4.1, cost between $0.3$ and $1.2$ US cents per \mantra test case (see Table~\ref{tab:ours-labeler-cost-short} in App.~\ref{app:cost-latency}),
whereas using a fine-tuned open-weights labeler incurs zero cost per test case.
Regarding latency, both our shields' and 2 of 3 baselines' overhead over not using a shield is negligible (see Table~\ref{tab:latency-per-call} in App.~\ref{app:cost-latency}).

\providecommand{\passAtOne}{Pass@1}
\providecommand{\passAtK}[1]{Pass@#1}
\providecommand{\passSupK}[1]{Pass$^{#1}$}

\providecommand{\notrun}{\multicolumn{1}{c}{--}}
\colorlet{propbg}{black!8}

\section{Discussion, Limitations and Future work}

Here, we discuss the limitations of \framework and propose directions for future work.

\paragraph{Autoformalization.}
\framework requires the natural language manual to be translated into formal specifications.
This process is difficult both for humans and LLMs, and our loop requires strong LLMs to produce satisfactory results and does not guarantee correctness, especially for ambiguous rules that may require careful interpretation.
Investigating better autoformalization techniques is a promising direction for future work that will improve \framework.

\paragraph{Labeler.} \framework requires a labeler LLM to ground the natural-language entities mentioned in the task prompt and tool outputs into formal variables to be used by the shield.
While we saw that this is a simple task for modern LLMs, even small finetuned open-weights models, it still introduces a layer of variability to the shield.
It would be interesting to explore how well deterministic labelling techniques can replace the labeler LLM.

\paragraph{Natural language.}
\framework acts only on the tool calls the agents generate, and cannot ensure correctness of their natural language outputs. As such, it is not applicable to settings requiring interaction with humans (for instance,~\citet{barres2025tau}).
Extending our shield to such settings is an interesting avenue for future work.

\paragraph{Token shield.}
The token shield must be reimplemented for different LLMs, tokenizers, or tool-call harnesses, as it relies on the specific tool calling format used during generation.
Otherwise, it may harm performance by causing unexpected token generations.

\ifdefined\psfW\else\newlength\psfW\newlength\psfL\fi
\begin{table}[t]
\centering
\fontsize{7pt}{8pt}\selectfont   %
\setlength{\tabcolsep}{2.5pt}
\setlength{\aboverulesep}{0pt}\setlength{\belowrulesep}{0pt}   %
\def\psfA{\rule{0pt}{7.6pt}}\def\psfB{\rule[-3.4pt]{0pt}{3.4pt}}   %
\settowidth{\psfW}{\textbf{00$\pm$0}}
\settowidth{\psfL}{GPT-5.4 mini}
\ifdim\psfL>\dimexpr2\psfW+2\tabcolsep\relax\setlength{\psfW}{\dimexpr(\psfL-2\tabcolsep)/2\relax}\fi
\definecolor{psfOurs}{HTML}{E6F3F1}   %
\makeatletter   %
\def\psfEdgeL{\gdef\CT@cell@color{\@tempdimb\z@\global\let\CT@cell@color\relax}}
\def\psfEdgeR{\gdef\CT@cell@color{\@tempdimc\z@\global\let\CT@cell@color\relax}}
\makeatother
\settowidth{\psfL}{\begin{tabular}{@{}l@{}}(a) \mantra\\(b) \mantra\\Ours (prompt)\\\textsc{AgentSpec}\\\textsc{Progent}\\\textsc{DynaGuard}\end{tabular}}
\noindent Agents: L3.1~= Llama~3.1, Q3~= Qwen3, Q3.6~= Qwen3.6, G4~= Gemma~4\par\vspace{2pt}
\begin{tabular}{@{}p{\psfL}*{4}{>{\centering\arraybackslash}p{\psfW}}!{\hspace{1pt}{\color{black!30}\vrule}\hspace{1pt}}*{4}{>{\centering\arraybackslash}p{\psfW}}!{\hspace{1pt}{\color{black!30}\vrule}\hspace{1pt}}*{4}{>{\centering\arraybackslash}p{\psfW}}!{\hspace{1pt}{\color{black!30}\vrule}\hspace{1pt}}*{4}{>{\centering\arraybackslash}p{\psfW}}@{}}
\toprule
\psfA (a) \mantra & \multicolumn{4}{c!{\hspace{1pt}{\color{black!30}\vrule}\hspace{1pt}}}{Operations ($N{=}47$)} & \multicolumn{4}{c!{\hspace{1pt}{\color{black!30}\vrule}\hspace{1pt}}}{Sneaky Sasquatch ($N{=}43$)} & \multicolumn{4}{c!{\hspace{1pt}{\color{black!30}\vrule}\hspace{1pt}}}{$\tau^2$-retail ($N{=}57$)} & \multicolumn{4}{c@{}}{$\tau^2$-airline ($N{=}44$)} \\
Guard\psfB & L3.1 & Q3 & Q3.6 & G4 & L3.1 & Q3 & Q3.6 & G4 & L3.1 & Q3 & Q3.6 & G4 & L3.1 & Q3 & Q3.6 & G4 \\
\midrule
\psfA None & \phantom{0}6$\pm$4 & 45$\pm$7 & 60$\pm$7 & 51$\pm$7 & 21$\pm$6 & 33$\pm$7 & 53$\pm$8 & 23$\pm$7 & \phantom{0}5$\pm$3 & 11$\pm$4 & 21$\pm$5 & 16$\pm$5 & 20$\pm$6 & 36$\pm$7 & 52$\pm$8 & 34$\pm$7 \\
\psfA \textsc{AgentSpec} & 13$\pm$5 & 55$\pm$7 & \textbf{72$\pm$7} & 62$\pm$7 & 14$\pm$5 & 28$\pm$7 & 63$\pm$7 & 16$\pm$6 & \phantom{0}9$\pm$4 & 25$\pm$6 & 26$\pm$6 & 19$\pm$5 & 30$\pm$7 & 43$\pm$8 & 50$\pm$8 & 43$\pm$8 \\
\textsc{Progent} & \phantom{0}4$\pm$3 & 47$\pm$7 & 68$\pm$7 & 47$\pm$7 & 21$\pm$6 & 30$\pm$7 & 51$\pm$8 & 23$\pm$7 & \phantom{0}9$\pm$4 & 12$\pm$4 & 25$\pm$6 & 21$\pm$5 & 27$\pm$7 & 39$\pm$7 & 45$\pm$8 & 39$\pm$7 \\
\textsc{DynaGuard} & 11$\pm$5 & 21$\pm$6 & 40$\pm$7 & 28$\pm$7 & 19$\pm$6 & 30$\pm$7 & 49$\pm$8 & 19$\pm$6 & 12$\pm$4 & 23$\pm$6 & 19$\pm$5 & 19$\pm$5 & 34$\pm$7 & 34$\pm$7 & 39$\pm$7 & 27$\pm$7 \\
\rowcolor{psfOurs}[\dimexpr\tabcolsep+1pt\relax][\dimexpr\tabcolsep+1pt\relax]\psfA \psfEdgeL Ours (prompt) & 36$\pm$7 & \textbf{72$\pm$7} & 66$\pm$7 & \textbf{72$\pm$7} & 35$\pm$7 & \textbf{63$\pm$7} & \textbf{70$\pm$7} & \textbf{65$\pm$7} & 16$\pm$5 & 33$\pm$6 & \textbf{33$\pm$6} & \textbf{40$\pm$7} & 41$\pm$7 & \textbf{64$\pm$7} & \textbf{64$\pm$7} & \psfEdgeR \textbf{50$\pm$8} \\
\rowcolor{psfOurs}[\dimexpr\tabcolsep+1pt\relax][\dimexpr\tabcolsep+1pt\relax]\psfEdgeL Ours (token)\psfB & \textbf{38$\pm$7} & 55$\pm$7 & 66$\pm$7 & 57$\pm$7 & \textbf{42$\pm$8} & 37$\pm$7 & 58$\pm$8 & \textbf{65$\pm$7} & \textbf{30$\pm$6} & \textbf{40$\pm$7} & 18$\pm$5 & 25$\pm$6 & \textbf{57$\pm$8} & 61$\pm$7 & 55$\pm$8 & \psfEdgeR \textbf{50$\pm$8} \\
\bottomrule
\end{tabular}

\vspace{7pt}
\begin{tabular}{@{}p{\psfL}*{2}{>{\centering\arraybackslash}p{\dimexpr2\psfW+2\tabcolsep\relax}}!{\hspace{1pt}{\color{black!30}\vrule}\hspace{1pt}}*{2}{>{\centering\arraybackslash}p{\dimexpr2\psfW+2\tabcolsep\relax}}!{\hspace{1pt}{\color{black!30}\vrule}\hspace{1pt}}*{2}{>{\centering\arraybackslash}p{\dimexpr2\psfW+2\tabcolsep\relax}}!{\hspace{1pt}{\color{black!30}\vrule}\hspace{1pt}}*{2}{>{\centering\arraybackslash}p{\dimexpr2\psfW+2\tabcolsep\relax}}@{}}
\toprule
\psfA (b) \mantra & \multicolumn{2}{c!{\hspace{1pt}{\color{black!30}\vrule}\hspace{1pt}}}{Operations} & \multicolumn{2}{c!{\hspace{1pt}{\color{black!30}\vrule}\hspace{1pt}}}{Sneaky Sasquatch} & \multicolumn{2}{c!{\hspace{1pt}{\color{black!30}\vrule}\hspace{1pt}}}{$\tau^2$-retail} & \multicolumn{2}{c@{}}{$\tau^2$-airline} \\
Guard\psfB & GPT-5.4 mini & Haiku 4.5 & GPT-5.4 mini & Haiku 4.5 & GPT-5.4 mini & Haiku 4.5 & GPT-5.4 mini & Haiku 4.5 \\
\midrule
\psfA None & 62$\pm$7 & 64$\pm$7 & 49$\pm$8 & 58$\pm$8 & 21$\pm$5 & 23$\pm$6 & 34$\pm$7 & 55$\pm$8 \\
\rowcolor{psfOurs}[\dimexpr\tabcolsep+1pt\relax][\dimexpr\tabcolsep+1pt\relax]\psfEdgeL Ours (prompt)\psfB & \textbf{77$\pm$6} & \textbf{72$\pm$7} & \textbf{63$\pm$7} & \textbf{72$\pm$7} & \textbf{40$\pm$7} & \textbf{26$\pm$6} & \textbf{61$\pm$7} & \psfEdgeR \textbf{64$\pm$7} \\
\bottomrule
\end{tabular}

\vspace{7pt}
\begin{tabular}{@{}l*{4}{>{\centering\arraybackslash}p{\psfW}}!{\hspace{1pt}{\color{black!30}\vrule}\hspace{1pt}}*{4}{>{\centering\arraybackslash}p{\psfW}}@{}}
\toprule
\psfA (c) \textsc{AgentOrca} & \multicolumn{4}{c!{\hspace{1pt}{\color{black!30}\vrule}\hspace{1pt}}}{Bank ($N{=}51$)} & \multicolumn{4}{c@{}}{Healthcare ($N{=}39$)} \\
Guard\psfB & L3.1 & Q3 & Q3.6 & G4 & L3.1 & Q3 & Q3.6 & G4 \\
\midrule
\psfA None & \phantom{0}0$\pm$0 & 37$\pm$7 & 82$\pm$5 & 33$\pm$7 & \phantom{0}0$\pm$0 & 23$\pm$7 & 87$\pm$5 & 15$\pm$6 \\
\rowcolor{psfOurs}[\dimexpr\tabcolsep+1pt\relax][\dimexpr\tabcolsep+1pt\relax]\psfA \psfEdgeL Ours (prompt) & \textbf{76$\pm$6} & \textbf{80$\pm$6} & \textbf{90$\pm$4} & \textbf{80$\pm$6} & \textbf{85$\pm$6} & \textbf{92$\pm$4} & \textbf{92$\pm$4} & \psfEdgeR \textbf{82$\pm$6} \\
\rowcolor{psfOurs}[\dimexpr\tabcolsep+1pt\relax][\dimexpr\tabcolsep+1pt\relax]\psfEdgeL Ours (token)\psfB & 43$\pm$7 & 63$\pm$7 & 84$\pm$5 & 67$\pm$7 & 59$\pm$8 & 79$\pm$7 & 90$\pm$5 & \psfEdgeR 74$\pm$7 \\
\bottomrule
\end{tabular}
\caption{\passSupFive{}~$\pm$~standard error (\%) over the $N$ test cases of a domain. (a)~Open-weights agents on \mantra against the baselines. (b)~Proprietary agents on \mantra domains. (c)~Open-weights agents on \textsc{AgentOrca}. None denotes no shield, and the highest value in each column is bolded.}
\label{tab:passsup5-by-domain}
\end{table}

\section{Conclusion}\label{sec:conclusion}
We present \framework, a runtime shield for tool-using LLM agents to guide them through complex workflows described in procedural manuals.
By combining strategy templates from temporal requirements with argument constraints encoded in SMT, \framework can act during inference to block unsafe tool calls that violate the manual and advise LLMs on how to make progress towards their assigned tasks.
Using \framework on several open-weights and proprietary agents, we show that it effectively boosts their performance and makes them more reliable in compliance-critical settings.

\newpage

\label{beforebibliography}

\newoutputstream{pages}
\openoutputfile{main.pg}{pages}
\addtostream{pages}{\getpagerefnumber{beforebibliography}}
\closeoutputstream{pages}

\bibliography{iclr2027_conference}
\bibliographystyle{iclr2027_conference}

\appendix

\clearpage

\section{Extended Related Work}\label{app:related-work}
\DeclareRobustCommand{\hbfull}{\tikz[baseline=-0.55ex]\fill (0,0) circle (0.8ex);}
\DeclareRobustCommand{\hbhalf}{\tikz[baseline=-0.55ex]{\fill (0,0) -- (90:0.8ex) arc (90:270:0.8ex) -- cycle; \draw (0,0) circle (0.8ex);}}
\DeclareRobustCommand{\hbnone}{\tikz[baseline=-0.55ex]\draw (0,0) circle (0.8ex);}
\DeclareRobustCommand{\hbna}{\textcolor{black!50}{--}}

\subsection{Comparison of Shielding Features}
We summarize the discussion of related shielding approaches for LLM agents from Sec.~\ref{sec:introduction} in Tab.~\ref{tab:shield-comparison} using the following criteria:
\begin{description}%
\item[Auto-extracted:] \hbfull{} if an LLM produces the checked constraints without hand-written rules, \hbhalf{} LLM-assisted constraint creation, \hbna{} if the shield checks no explicit constraints.
\item[Deterministic:] \hbfull{} if the compliance decisions are deterministic (without an LLM judge), \hbhalf{} if an LLM edits constraints during the run. 
\item[Temporal order, argument values:] \hbhalf{} if left to an LLM judge or hand-coded as general-purpose code.
\item[Obligations:] \hbfull{} if the shield tracks open duties and does not let the agent end the task while one is open, \hbhalf{} if open duties can only be detected at the end (by hand-coded checks or an error flag).
\item[Free-text outputs:] \hbfull{} if the shield can decide based on natural-language tool results, \hbhalf{} if this needs hand-written string checks or only changes the policy indirectly.
\item[Guidance:] \hbfull{} if a blocked call is answered with a compliant alternative or the action that unlocks it, \hbhalf{} if only the reason for the block is given.
\item[Token-level:] enforcement during decoding, for open-source models.
\end{description}

\begin{table}[h]
\centering
\caption{Comparison of \framework with existing LLM guarding approaches: \hbfull{} supported, \hbhalf{} partially supported, \hbnone{} not supported, \hbna{} not applicable.}
\label{tab:shield-comparison}
\footnotesize
\setlength{\tabcolsep}{3.5pt}
\renewcommand{\arraystretch}{1.25}
\resizebox{\textwidth}{!}{%
\begin{tabular}{@{}l c ccccc cc@{}}
\toprule
& Specification & \multicolumn{5}{c}{Runtime check} & \multicolumn{2}{c}{Agent support} \\
\cmidrule(lr){2-2} \cmidrule(lr){3-7} \cmidrule(l){8-9}
Shield
& \shortstack{Auto-\\extracted}
& \shortstack{Determ-\\inistic} & \shortstack{Temporal\\order} & \shortstack{Argument\\values} & \shortstack{Oblig-\\ations} & \shortstack{Free-text\\outputs}
& Guidance & \shortstack{Token-\\level} \\
\midrule
\multicolumn{9}{@{}l}{\textit{LLM-based}} \\
\quad GuardAgent~\citep{xiang2024guardagent} & \hbfull & \hbnone & \hbnone & \hbhalf & \hbnone & \hbfull & \hbhalf & \hbnone \\
\quad DynaGuard~\citep{hoover2026dynaguard}  & \hbna   & \hbnone & \hbhalf & \hbhalf & \hbnone & \hbfull & \hbhalf & \hbnone \\
\quad PolicyGuide~\citep{kang2026policyguide} & \hbfull & \hbnone & \hbhalf & \hbhalf & \hbhalf & \hbfull & \hbfull & \hbnone \\
\multicolumn{9}{@{}l}{\textit{Rule-based}} \\
\quad AgentSpec~\citep{wang2025agentspec}    & \hbhalf & \hbfull & \hbhalf & \hbhalf & \hbhalf & \hbhalf & \hbhalf & \hbnone \\
\quad Progent~\citep{shi2025progent}         & \hbhalf & \hbhalf & \hbnone & \hbfull & \hbnone & \hbhalf & \hbhalf & \hbnone \\
\quad Agent-C~\citep{kamath2025enforcing}    & \hbhalf & \hbfull & \hbfull & \hbfull & \hbhalf & \hbhalf & \hbhalf & \hbfull \\
\midrule[\heavyrulewidth]
\textbf{\framework{} (ours)}           & \hbfull & \hbfull & \hbfull & \hbfull & \hbfull & \hbfull & \hbfull & \hbfull \\
\bottomrule
\end{tabular}%
}
\end{table}

Tab.~\ref{tab:shield-comparison} shows that \framework covers all criteria, while every existing technique lacks several of them. Furthermore, PolicyGuide and Agent-C have not yet released their code, so we cannot compare against them in our experiments.
In the following, we show how the lack of these features results in non-compliant traces that slip through these shields, as observed in our experiments.

\clearpage
\subsection{Failure Traces of other Shielding Techniques w.r.t.\ \chal{1}--\chal{2}}
\label{app:failure-traces-baselines}

This section describes how non-compliant traces in our experiments slip through existing shields, and why these shields often prevented the agent from completing its task at all.
We group the observed failures by their cause.

\paragraph{No guidance, so the agent stalls (\chal{1}, guidance).}
The most common failure mode of other shields that we observed is \emph{stalling}, which manifests when the shield blocks a call but does not offer any guidance to the agent.
As a consequence, the agent repeats the blocked call, or tries other blocked calls until it exhausts its budget without completing the task.

As our example, consider the business class booking task from Figure~\ref{fig:shield-example}.
Here, the user asks for a business class 1.5h flight within North America.
This violates the manual because it forbids business class on flights shorter than 6h unless there is an urgent business need.
Since the user does not state an urgent business need, the agent must instead book an economy flight.

\textsc{AgentSpec} blocks the request to book business class in 13 of 20 runs on this task, but does not tell the agent what to do instead,
so 5 runs fail to book any flight at all.

\textsc{DynaGuard}'s LLM judge explains why a call was blocked, but only sees prior executed calls on the history and not previously blocked ones.
As a result, the agents often repeat the same (blocked) call multiple times and receive the same explanation from the judge.
Further, the LLM judge itself often misrepresents the manual.
We observed this kind of stalling behavior in multiple runs and tasks, and show a different flight booking example in Figure~\ref{fig:stalling} on a task from the $\tau^2$-airline domain in \mantra.
Here, \textsc{DynaGuard} blocks 50 calls of Gemma~4 in a row, claiming that the agent must ask the user for confirmation before doing anything.
However, on the manual, this is only required for specific tasks, so the judge's verdict is wrong and the agent has no way to bypass the blocks.

\framework, in contrast, suggests an exact tool call that would be allowed (either a prerequisite to the blocked call or a repaired version of the blocked call) so the agent always has access to a compliant call that makes progress.
In the business class booking example, \framework suggests calling \texttt{book\_flight} with the argument \texttt{travel\_class=economy}.

\paragraph{No access to the history (\chal{1}, context-dependence).}
Whether a tool call complies may depend on the history of the task.
To illustrate this, consider a testcase from the Operations domain of \mantra, where the user asks the agent to install a software.
The intended workflow according to the manual is to first call \texttt{check\_software\_tier} and call \texttt{run\_vulnerability\_scan}.
Depending on the results of those calls it is determined whether it is safe to proceed with the call \texttt{install\_software}, and finally, before ending the task, the agent must call \code{archive\_scan\_report}.
Since whether \texttt{install\_software} is allowed depends on results of earlier calls, a shield must take this history into account to allow or block the call accordingly during the task execution.

\textsc{Progent} fails because its rules check each call only against its own arguments and do not refer to earlier calls or their results.
The only memory is an LLM that may rewrite the rules after each allowed call.
However, in our experiments, we observed that these rewrites failed to capture this context-dependence.
Over all runs and testcases in this domain, only once did the same call change from blocked to allowed, as \texttt{install\_software} should be here.

On the other hand, \textsc{AgentSpec} can use the history, since its rules read the earlier calls and their results.
In this testcase, one rule correctly blocks \texttt{install\_software} until \texttt{check\_software\_tier} has returned a result for that software.

Lastly, \textsc{DynaGuard} uses an LLM judge that can read the history of executed calls before giving a verdict,
but its reliability varies unpredictably.
On this testcase, in one run it correctly blocked \texttt{install\_software} because no vulnerability scan was run and later allowed it after the agent ran the scan.
However, in another run, it failed to recognize that the agent had successfully completed the workflow and had run \texttt{archive\_scan\_report}, and as a result kept blocking \texttt{done} and insisting that the archival is still pending.

In contrast, \framework's strategy template both keeps track of open duties and captures context-dependent constraints.

\begin{figure}[t]
\centering
\definecolor{stallRed}{HTML}{C0392B}
\definecolor{stallMuted}{HTML}{6B6B6B}
\setlength{\fboxsep}{3pt}\setlength{\fboxrule}{0.4pt}
\fcolorbox{stallMuted}{white}{\begin{minipage}{\dimexpr0.55\textwidth-2\fboxsep-2\fboxrule}
\fontsize{6}{7.2}\selectfont\ttfamily\setlength{\parskip}{0pt}\setlength{\parindent}{0pt}
\textcolor{stallMuted}{Task: for U-88104, book a one-way economy trip RDU -> BOI on 2026-10-03; reuse passenger, bags and payment of reservation R-551208.}\par
\vspace{4pt}
\begin{tabular}{@{}r@{\ }l@{\ \ }l@{}}
{[}0{]}     & get\_reservation\_details(R-551208)   & \textcolor{stallRed}{BLOCKED} \\
{[}1{]}     & search\_direct\_flight(10-03,RDU,BOI)  & \textcolor{stallRed}{BLOCKED} \\
{[}2{]}     & search\_onestop\_flight(10-03,RDU,BOI) & \textcolor{stallRed}{BLOCKED} \\
{[}3{]}     & search\_direct\_flight(...)            & \textcolor{stallRed}{BLOCKED} \\
{[}4{]}     & get\_reservation\_details(...)         & \textcolor{stallRed}{BLOCKED} \\
{[}5-44{]}  & the same three calls round-robin       & \textcolor{stallRed}{BLOCKED} \\
{[}45{]}    & search\_direct\_flight(...)            & \textcolor{stallRed}{BLOCKED} \\
            & \multicolumn{2}{@{}l@{}}{[HARNESS] 45 tool calls made; call done now.} \\
{[}46-49{]} & the same three calls round-robin       & \textcolor{stallRed}{BLOCKED} \\
            & \multicolumn{2}{@{}l@{}}{Step cap reached: no book\_reservation, no done.} \\
\end{tabular}\par
\vspace{4pt}
\textcolor{stallMuted}{Guard's reason at step 1: "The agent violated the rule by not listing the action details and obtaining explicit user confirmation before making a tool call. The rule states 'Before taking any actions that update the booking database (...) you must list the action details and obtain explicit user confirmation'."}\par
\end{minipage}}
\caption{Agent stalling: \textsc{DynaGuard} on Gemma~4 executing a testcase from the $\tau^2$-airline domain in \mantra.
All 50 calls, including read-only lookups, are blocked with the same (sometimes incorrect) reason.}
\label{fig:stalling}
\end{figure}

\paragraph{No argument constraints (\chal{2}).}
Many tasks require constraints on the arguments of a call, not only its order in the trace.
We return to the business class flight example to illustrate this: economy class is allowed and business is not because there is no urgent business need.
If the agent attempts to book business class, the tool output itself returns a rejection and indicates that an urgent business need is required.
Without any shield, the agents often subsequently invent fake such urgent business needs in attempt to book business class.

\textsc{Progent}'s rules only check argument types and not exact arguments, e.g., \texttt{travel\_class} must be economy, business or first.
As a consequence, \textsc{Progent} does not block the business class requests and does not catch the fake urgent business needs.

\textsc{DynaGuard}'s LLM judge is, again, inconsistent.
In 12 runs, the initial business class request is blocked, but in 2 others it is allowed. 
The fake urgent business need is also blocked in 10 runs but allowed in 2 others.

\textsc{AgentSpec} successfully blocks the initial business request in 13 of 20 runs, but even then, as discussed above, does not guide towards booking economy.

\framework, in contrast, blocks the initial business request and suggests the same call with \texttt{travel\_class="economy"}, so the agent does not receive the suggestion to give an urgent business need.
In the one run where the agent still happened to write an urgent business need, our shield blocks it, because this is allowed only if the user stated an urgent business need in their request.

\clearpage
\section{Specification Language}\label{app:spec-language}

This section gives the syntax and semantics of the specification language introduced in Sec.~\ref{sec:spec}. A specification contains temporal requirements and
argument constraints, together with the declarations that connect them to tool calls and their outcomes. Figure~\ref{fig:spec-grammar} summarizes its structure.

Requirements are parameterized by concrete entities.
An entity type specifies how tool arguments and outcomes identify entities and a call may identify several entities.
Requirements concerning the same entity type are grouped into a \emph{module}, which declares its propositions, temporal requirements, and argument constraints.
Several modules may concern the same entity type. Within a module, the parameter $v$ denotes the entity being considered. It may represent a tuple, such as
the route $(x,y)$ in Sec.~\ref{sec:spec}. Requirements concerning the whole task use a common task identifier.

Each entity-identification rule associates a tool with an expression that computes an entity key from the tool's arguments and, when available, its outcome. The rule is evaluated on proposed arguments before the call is checked and again on the completed event after execution. A scalar result identifies one entity, a collection identifies each of its elements, and an undefined result identifies no entity. Entity keys are compared after conversion to strings, removal of surrounding
whitespace, and case folding. Tuple-valued parameters are represented by a composite key whose components preserve the corresponding argument values.

The specification also declares the information to be extracted from natural language: schemas for pseudo tool calls representing requests, and label vocabularies for free-text tool outcomes. These declarations determine the labeler's inputs and outputs (Sec.~\ref{sec:online}). Each temporal requirement and argument constraint is
associated with its source sentence in the manual, which can subsequently be included in the shield's feedback.

\newcommand{\snt}[1]{\ensuremath{\langle\mathit{#1}\rangle}}

\begin{figure}[t]
\centering
\small
\begin{tabular}{@{}r@{\;}c@{\;}l@{}}
\multicolumn{3}{@{}l}{
  \textit{(a) Specification structure}}\\
\snt{spec}
& ::= &
$(\{\snt{type}\},\{\snt{module}\},
  \{\snt{request}\},\{\snt{observation}\})$\\
\snt{type}
& ::= & $(S,\kappa)$\\
\snt{module}
& ::= &
$(S,\{\snt{proposition}\},
  \{\snt{requirement}\},\{\snt{constraint}\})$\\
\snt{proposition}
& ::= & $(p(v),t,e)$\\
\snt{requirement}
& ::= & $(\mathsf{kind},\varphi,m)$\\
\snt{constraint}
& ::= & $(p(v),e,m)$\\
\snt{request}
& ::= & $(Q,\Sigma,d,I)$\\
\snt{observation}
& ::= & $(t,\pi_{\mathrm{text}},
             \pi_{\mathrm{label}},\Lambda,I)$\\[4pt]

\multicolumn{3}{@{}l}{
  \textit{(b) Temporal requirements via LTL}}\\
$\varphi$
& ::= &
$\top \mid \bot \mid p(v) \mid \ltlO p(v)
 \mid \ltlnot\varphi
 \mid (\varphi\ltland\varphi)
 \mid (\varphi\ltlor\varphi)
 \mid (\varphi\ltlimplies\varphi)$\\
& $\mid$ &
$\ltlX\varphi \mid \ltlF\varphi \mid \ltlG\varphi
 \mid (\varphi\ltlU\varphi)
 \mid (\varphi\ltlW\varphi)$\\[4pt]

\multicolumn{3}{@{}l}{
  \textit{(c) Core expressions for proposition predicates
  and argument constraints}}\\
$e$
& ::= &
$c \mid v \mid a \mid o \mid b_v
 \mid e.\pi \mid e[e]
 \mid [e_1,\ldots,e_r]$\\
& $\mid$ &
$\ltlnot e \mid -e
 \mid (e\ltland e) \mid (e\ltlor e)
 \mid (e\bowtie e) \mid (e\odot e)
 \mid \operatorname{if}(e,e,e)$\\
& $\mid$ &
$\operatorname{last}(p(v))
 \mid \operatorname{exists}(p(v))
 \mid \operatorname{count}(p(v))
 \mid \operatorname{hist}(p(v),\pi)$\\
& $\mid$ &
$\operatorname{present}(e)
 \mid \operatorname{len}(e)
 \mid \operatorname{contains}(e,e)
 \mid \operatorname{startswith}(e,e)$\\
$\bowtie$
& $\in$ &
$\{\code{==},\code{!=},<,\leq,>,\geq,\in,\notin\}$\\
$\odot$
& $\in$ & $\{+,-,\times,/\}$\\
\end{tabular}
\caption{
Abstract structure and syntax of the specification
language. Braces denote finite collections.
In (a), $S$ is an entity type and $\kappa$ its
entity-identification rules; $\mathsf{kind}$ classifies a
temporal requirement as a safety requirement or an
obligation, and $m$ is its source sentence.
A request declaration gives a pseudo tool name $Q$,
a typed argument schema $\Sigma$, defaults $d$,
and extraction instructions $I$.
An observation declaration identifies the free-text
and label fields of an outcome and its label vocabulary
$\Lambda$. In (c), $c$ is a constant, $\pi$ a field
path, and $b_v$ the event introducing the entity.
The function meanings are given in
Table~\ref{tab:spec-functions}.
Auxiliary string, list, and conversion helpers are
omitted from the core expression grammar.
}
\label{fig:spec-grammar}
\end{figure}

\paragraph{Temporal Requirements via LTL.}
A proposition declaration $(p(v),t,e)$ associates $p(v)$ with tool $t$ and a Boolean predicate $e$. The proposition holds at an event precisely when the event calls $t$ and $e$ evaluates to true under the entity binding and an omitted predicate is interpreted as true. Matching the event to the entity must be expressed in $e$, as the parameter $v$ alone does not restrict a proposition to calls concerning that entity.
For example, a proposition for route $(x,y)$ compares the call's origin and destination with $x$ and $y$.

An \emph{action proposition} can be evaluated from the proposed call, the entity binding, and recorded history. An \emph{observation proposition} additionally
depends on the current outcome. Earlier outcomes are available through the history,
whereas the current outcome becomes available only after execution. This distinction determines which requirements can be checked before a tool call.

Fix a module and an entity binding, and let $\mathcal{P}$ be the resulting set of propositions. Each recorded event induces a letter $\ell_i\subseteq\mathcal{P}$ containing precisely the propositions that hold at that event. The resulting
finite word is $\sigma=\ell_0\cdots\ell_{n-1}\in(2^{\mathcal{P}})^+$. Events at which no proposition holds retain their positions and contribute the empty letter.
Formulas follow standard LTL syntax (see~Fig.~\ref{fig:spec-grammar}(b)). Using finite-trace semantics~\citep{de2013linear}, extended with the past operator $\ltlO$, satisfaction at position $0\leq i<n$ is defined by
\[
\begin{aligned}
\sigma,i\models p
&\iff p\in\ell_i,\\
\sigma,i\models\ltlnot\varphi
&\iff \sigma,i\not\models\varphi,\\
\sigma,i\models\varphi\ltland\psi
&\iff (\sigma,i\models\varphi)
       \land(\sigma,i\models\psi),\\
\sigma,i\models\ltlX\varphi
&\iff i+1<n
       \land(\sigma,i+1\models\varphi),\\
\sigma,i\models\varphi\ltlU\psi
&\iff \exists j,\ i\leq j<n:
       (\sigma,j\models\psi)\\
&\qquad{}\land
       \bigl(\forall k,\ i\leq k<j:
       \sigma,k\models\varphi\bigr),\\
\sigma,i\models\ltlO p
&\iff \exists j,\ 0\leq j\leq i:\ p\in\ell_j.
\end{aligned}
\]
Here, the bound entity parameter is suppressed in proposition names. Boolean constants, disjunction, and implication have their usual meanings, and
\[
\begin{aligned}
\ltlF\varphi
&\equiv \top\ltlU\varphi,
&
\ltlG\varphi
&\equiv \ltlnot\ltlF\ltlnot\varphi,\\
\varphi\ltlW\psi
&\equiv(\varphi\ltlU\psi)\ltlor\ltlG\varphi.
\end{aligned}
\]
We write $\sigma\models\varphi$ iff
$\sigma,0\models\varphi$.
In particular, $\ltlX\varphi$ is false at the final position as it requires a further event.

The past operator $\ltlO$ expresses that a proposition holds at the current event or held earlier. It permits prerequisites to be written directly without increasing the expressive power of LTL~\citep{gabbay1980temporal,de2020pure}. 

For example, with $p$ denoting a distance check and $q$ a booking for the same route,
$\ltlG(q\ltlimplies\ltlO p)$ expresses the prerequisite in P1. Since checking and booking are distinct tool calls, the check must precede the booking. Obligations may instead require a successful outcome, for example a failed booking attempt does not satisfy
$\ltlF\,\ltlprop{booked\_flight\_ok}(x,y)$. An activated obligation that remains unfulfilled is an \emph{open duty}, as in Sec.~\ref{sec:spec}. If the manual permits the work to be waived under a condition $r$, that alternative must be explicit in the formula, for example $\ltlF(q\ltlor r)$. A failed tool call does not by itself discharge an obligation.

\paragraph{Argument Constraints via SMT.}
An argument constraint associates an action proposition $p(v)$ with a Boolean expression $e$. Whenever a proposed call satisfies $p(v)$, its arguments must satisfy $e$ under the current entity binding and recorded history. The condition may use outcomes of earlier calls, but cannot depend on the unavailable outcome of
the proposed call. This allows constraints to express, for example, that an identifier was returned by an earlier lookup, that a required argument is present, or that an argument value is permitted by previously established facts.

The DSL in Fig.~\ref{fig:spec-grammar}(c) provides field and list access, Boolean and arithmetic expressions, comparisons, conditionals, and operations
on recorded history. Here, $a$ and $o$ denote the current call's arguments and outcome, $v$ the bound entity, and $b_v$ the event that introduced it. Access to $o$ is restricted to predicates evaluated after execution. History operations inspect earlier executed events recorded for the entity and module, excluding the current event, and are initialized by the labeler LLM.
Table~\ref{tab:spec-functions} defines the functions shown in the grammar.

\begin{table}[t]
\centering
\small
\caption{
Functions in the core expression language.
History operations use earlier recorded events with
the entity parameter $v$ bound.
A field path $\pi$ selects information from an event's
tool arguments or outcome.
}
\label{tab:spec-functions}
\begin{tabularx}{\textwidth}{@{}lX@{}}
\toprule
Expression & Meaning\\
\midrule
$\operatorname{last}(p(v))$
&
The latest earlier event satisfying $p(v)$;
if none exists, its fields are undefined
($\mathsf{null}$).\\
$\operatorname{exists}(p(v))$,
$\operatorname{count}(p(v))$
&
Whether an earlier event satisfies $p(v)$,
and the number of such events, respectively.\\
$\operatorname{hist}(p(v),\pi)$
&
The non-null values at path $\pi$ from earlier
events satisfying $p(v)$, in chronological order.\\
$\operatorname{present}(e)$
&
Whether $e$ is neither null, an empty string,
nor an empty list.\\
$\operatorname{len}(e)$
&
The length of a string or list;
zero for an unsuitable value.\\
$\operatorname{contains}(e_1,e_2)$
&
For a list $e_1$, whether it contains an element
equal to $e_2$; for a string, a case-insensitive
substring test.\\
$\operatorname{startswith}(e_1,e_2)$
&
A case-insensitive string-prefix test.\\
$\operatorname{if}(e_1,e_2,e_3)$
&
$e_2$ when the Boolean condition $e_1$ holds,
and $e_3$ otherwise.\\
\bottomrule
\end{tabularx}
\end{table}

Expression evaluation accounts for incomplete tool outputs. A missing field or list element evaluates to $\mathsf{null}$.
When a rule requires an earlier lookup to exist,that prerequisite must be expressed explicitly, for example using $\operatorname{exists}(p(v))$. If an applicable constraint evaluates to false, the proposed arguments do not satisfy it.

Concrete proposed arguments are checked by evaluating these expressions. For symbolic argument search, the SMT encoding introduces variables for the arguments and evaluates
history-dependent subexpressions to fixed values. String, numeric, and Boolean arguments are represented by string, real, and Boolean variables, respectively. The supported Boolean, arithmetic, and string operations are then encoded using the corresponding SMT operations. This representation is used by Z3~\citep{demoura2008z3}
to find argument values satisfying the proposition predicates and applicable constraints at runtime.

For the running example, let $B_{x,y}=\code{business\_allowed}(h,x,y)$ be the history-dependent Boolean from Sec.~\ref{sec:spec}. It records both an urgent reason and the absence of an economy seat for the route. For a flight booking, the symbolic conditions expressing the entity binding and P3 are
\[
\begin{aligned}
&a[\code{origin}]=x
 \ \land\ a[\code{destination}]=y\\
&\qquad{}\land
 \bigl(a[\code{class}]\neq\code{business}
       \ \lor\ B_{x,y}\bigr).
\end{aligned}
\]
The route and $B_{x,y}$ are fixed by the current binding and history and the arguments are symbolic. For Toronto--Chicago in Fig.~\ref{fig:shield-example}, no urgent reason has been established, so $B_{x,y}$ is false. The resulting constraint excludes business class while preserving the route.

\clearpage
\section{Additional Experimental Results}\label{app:more-experiments}

\subsection{Additional Implementation Details}\label{app:setup}

\paragraph{Model and shield parameters.}
For \mantra, agents have a 40 tool call limit, and the shield stops blocking \texttt{done} after 25 tool calls regardless of if there still are open duties.
For SOPBench, these limits are 200 and 60 respectively, and for AgentOrca they are both 200.
The open-weights models are served by vLLM and the token shield is implemented via a logit processor.
We use default temperature, reasoning effort and top-$p$, except on the token shield where we use top-$p=1$ to avoid tokens being masked by the decoding.
Moreover, on the token shield, we set the boost strength defined in Sec.~\ref{sec:methodology} to $\eta = 0.5$. 
The full names of the open-weights LLM agents evaluated are Llama 3.1-8B-Instruct, Qwen-3-8B, Qwen3.6-27B and Gemma-4-E4B-Instruct.
The proprietary agents used for labelling or for evaluation are accessed via the OpenAI and Anthropic APIs, whereas the ones used for autoformalization are accessed via Codex and Claude Code.

\paragraph{Metric definitions.} For a benchmark suite with $N$ tasks, we run each task $n=5$ times independently and let $\alpha_i \in \{0, 1, \dots, n\}$ denote the number of successful runs on task $i$. The tables below report three standard success rates:
\begin{itemize}
	\item \passAtOne --- the average single-attempt success rate, $\tfrac{1}{N}\sum_{i=1}^{N} \tfrac{\alpha_i}{n}$. This is the standard accuracy of one independent rollout.
	\item \passAtK --- the probability that at least one of $k$ independent attempts on a task succeeds, averaged over the suite, $\text{Pass@}k = \tfrac{1}{N}\sum_{i=1}^{N}\!\left[1 - \binom{n - \alpha_i}{k}\big/\binom{n}{k}\right]$ for $n \geq k$. We report \passAtFive.
	\item \passSupK --- the probability that \emph{all} $k$ independent attempts on a task succeed, averaged over the suite~\citep{barres2025tau}: \passSupK $ = \tfrac{1}{N}\sum_{i=1}^{N} \binom{\alpha_i}{k}\big/\binom{n}{k}$. This is a stricter consistency/reliability measure than Pass@$k$. We report \passSupFive.
\end{itemize}

\clearpage
\subsection{Full Results on \mantra}\label{app:mantra-full}

\begin{table}[htbp]
\centering
\scriptsize
\setlength{\tabcolsep}{3pt}
\caption{\passAtOne{} on the four \mantra domains for open-weight (white) and proprietary (gray) agents.}
\label{tab:pass1-by-domain}
\begin{tabular}{@{}p{1.7cm}lrrrrrr@{}}
\toprule
Domain & Model & None & AgentSpec & Progent & DynaGuard & \multicolumn{1}{c}{Ours} & \multicolumn{1}{c@{}}{Ours} \\
 & & & & & & \multicolumn{1}{c}{(prompt)} & \multicolumn{1}{c@{}}{(token)} \\
\midrule
 & Llama 3.1 & 0.26$\pm$0.05 & 0.34$\pm$0.06 & 0.27$\pm$0.05 & 0.20$\pm$0.05 & \textbf{0.61$\pm$0.06} & 0.58$\pm$0.06 \\
 & Qwen3 & 0.62$\pm$0.06 & 0.71$\pm$0.06 & 0.66$\pm$0.06 & 0.34$\pm$0.06 & \textbf{0.80$\pm$0.05} & 0.70$\pm$0.06 \\
 & Qwen3.6 & 0.72$\pm$0.06 & \textbf{0.84$\pm$0.05} & 0.75$\pm$0.06 & 0.51$\pm$0.07 & 0.79$\pm$0.05 & 0.74$\pm$0.06 \\
 & Gemma 4 & 0.63$\pm$0.07 & 0.69$\pm$0.06 & 0.65$\pm$0.06 & 0.39$\pm$0.07 & \textbf{0.84$\pm$0.05} & 0.73$\pm$0.06 \\
\rowcolor{propbg}\cellcolor{white} & GPT-5.4 mini & 0.72$\pm$0.06 & \notrun & \notrun & \notrun & \textbf{0.82$\pm$0.05} & \notrun \\
\rowcolor{propbg}\cellcolor{white}\multirow{-6}{=}{Operations ($N=47$)} & Claude Haiku 4.5 & 0.76$\pm$0.05 & \notrun & \notrun & \notrun & \textbf{0.78$\pm$0.06} & \notrun \\
\midrule
 & Llama 3.1 & 0.48$\pm$0.06 & 0.40$\pm$0.06 & 0.46$\pm$0.06 & 0.46$\pm$0.06 & 0.59$\pm$0.06 & \textbf{0.61$\pm$0.06} \\
 & Qwen3 & 0.48$\pm$0.07 & 0.54$\pm$0.06 & 0.45$\pm$0.07 & 0.47$\pm$0.07 & \textbf{0.78$\pm$0.06} & 0.68$\pm$0.05 \\
 & Qwen3.6 & 0.67$\pm$0.07 & 0.66$\pm$0.07 & 0.61$\pm$0.07 & 0.63$\pm$0.07 & \textbf{0.77$\pm$0.06} & \textbf{0.77$\pm$0.05} \\
 & Gemma 4 & 0.41$\pm$0.06 & 0.31$\pm$0.06 & 0.40$\pm$0.06 & 0.36$\pm$0.06 & 0.73$\pm$0.06 & \textbf{0.74$\pm$0.06} \\
\rowcolor{propbg}\cellcolor{white} & GPT-5.4 mini & 0.71$\pm$0.05 & \notrun & \notrun & \notrun & \textbf{0.77$\pm$0.06} & \notrun \\
\rowcolor{propbg}\cellcolor{white}\multirow{-6}{=}{Sneaky Sasquatch ($N=43$)} & Claude Haiku 4.5 & 0.71$\pm$0.06 & \notrun & \notrun & \notrun & \textbf{0.80$\pm$0.06} & \notrun \\
\midrule
 & Llama 3.1 & 0.12$\pm$0.04 & 0.19$\pm$0.04 & 0.18$\pm$0.04 & 0.20$\pm$0.05 & 0.39$\pm$0.05 & \textbf{0.46$\pm$0.06} \\
 & Qwen3 & 0.31$\pm$0.05 & 0.47$\pm$0.05 & 0.33$\pm$0.05 & 0.31$\pm$0.06 & 0.53$\pm$0.06 & \textbf{0.57$\pm$0.06} \\
 & Qwen3.6 & 0.46$\pm$0.05 & \textbf{0.51$\pm$0.05} & 0.45$\pm$0.06 & 0.25$\pm$0.05 & 0.49$\pm$0.06 & 0.45$\pm$0.05 \\
 & Gemma 4 & 0.35$\pm$0.05 & 0.40$\pm$0.05 & 0.38$\pm$0.05 & 0.27$\pm$0.05 & \textbf{0.49$\pm$0.06} & 0.48$\pm$0.06 \\
\rowcolor{propbg}\cellcolor{white} & GPT-5.4 mini & 0.39$\pm$0.05 & \notrun & \notrun & \notrun & \textbf{0.53$\pm$0.06} & \notrun \\
\rowcolor{propbg}\cellcolor{white}\multirow{-6}{=}{$\tau^2$-retail ($N=57$)} & Claude Haiku 4.5 & 0.29$\pm$0.06 & \notrun & \notrun & \notrun & \textbf{0.41$\pm$0.06} & \notrun \\
\midrule
 & Llama 3.1 & 0.24$\pm$0.06 & 0.40$\pm$0.07 & 0.34$\pm$0.07 & 0.35$\pm$0.07 & 0.60$\pm$0.07 & \textbf{0.63$\pm$0.07} \\
 & Qwen3 & 0.46$\pm$0.07 & 0.51$\pm$0.07 & 0.45$\pm$0.07 & 0.37$\pm$0.07 & \textbf{0.75$\pm$0.06} & 0.70$\pm$0.07 \\
 & Qwen3.6 & 0.63$\pm$0.07 & 0.60$\pm$0.07 & 0.55$\pm$0.07 & 0.39$\pm$0.07 & \textbf{0.67$\pm$0.07} & 0.63$\pm$0.07 \\
 & Gemma 4 & 0.46$\pm$0.07 & 0.51$\pm$0.07 & 0.49$\pm$0.07 & 0.33$\pm$0.07 & \textbf{0.59$\pm$0.07} & 0.55$\pm$0.07 \\
\rowcolor{propbg}\cellcolor{white} & GPT-5.4 mini & 0.47$\pm$0.06 & \notrun & \notrun & \notrun & \textbf{0.65$\pm$0.07} & \notrun \\
\rowcolor{propbg}\cellcolor{white}\multirow{-6}{=}{$\tau^2$-airline ($N=44$)} & Claude Haiku 4.5 & \textbf{0.67$\pm$0.07} & \notrun & \notrun & \notrun & \textbf{0.67$\pm$0.07} & \notrun \\
\bottomrule
\end{tabular}
\end{table}

\begin{table}[htbp]
\centering
\scriptsize
\setlength{\tabcolsep}{3pt}
\caption{\passAtFive{} on the four \mantra domains for open-weight (white) and proprietary (gray) agents.}
\label{tab:pass5-by-domain}
\begin{tabular}{@{}p{1.7cm}lrrrrrr@{}}
\toprule
Domain & Model & None & AgentSpec & Progent & DynaGuard & \multicolumn{1}{c}{Ours} & \multicolumn{1}{c@{}}{Ours} \\
 & & & & & & \multicolumn{1}{c}{(prompt)} & \multicolumn{1}{c@{}}{(token)} \\
\midrule
 & Llama 3.1 & 0.43$\pm$0.07 & 0.60$\pm$0.07 & 0.47$\pm$0.07 & 0.32$\pm$0.07 & 0.77$\pm$0.06 & \textbf{0.81$\pm$0.06} \\
 & Qwen3 & 0.72$\pm$0.07 & 0.81$\pm$0.06 & 0.79$\pm$0.06 & 0.47$\pm$0.07 & \textbf{0.85$\pm$0.05} & 0.77$\pm$0.06 \\
 & Qwen3.6 & 0.79$\pm$0.06 & \textbf{0.89$\pm$0.05} & 0.81$\pm$0.06 & 0.64$\pm$0.07 & 0.85$\pm$0.05 & 0.85$\pm$0.05 \\
 & Gemma 4 & 0.70$\pm$0.07 & 0.74$\pm$0.06 & 0.79$\pm$0.06 & 0.49$\pm$0.07 & \textbf{0.87$\pm$0.05} & 0.83$\pm$0.06 \\
\rowcolor{propbg}\cellcolor{white} & GPT-5.4 mini & 0.81$\pm$0.06 & \notrun & \notrun & \notrun & \textbf{0.85$\pm$0.05} & \notrun \\
\rowcolor{propbg}\cellcolor{white}\multirow{-6}{=}{Operations ($N=47$)} & Claude Haiku 4.5 & \textbf{0.89$\pm$0.05} & \notrun & \notrun & \notrun & 0.85$\pm$0.05 & \notrun \\
\midrule
 & Llama 3.1 & 0.74$\pm$0.07 & 0.67$\pm$0.07 & 0.70$\pm$0.07 & 0.74$\pm$0.07 & \textbf{0.86$\pm$0.05} & 0.79$\pm$0.06 \\
 & Qwen3 & 0.60$\pm$0.08 & 0.72$\pm$0.07 & 0.58$\pm$0.08 & 0.60$\pm$0.08 & 0.86$\pm$0.05 & \textbf{0.91$\pm$0.04} \\
 & Qwen3.6 & 0.77$\pm$0.07 & 0.70$\pm$0.07 & 0.74$\pm$0.07 & 0.77$\pm$0.07 & 0.86$\pm$0.05 & \textbf{0.91$\pm$0.04} \\
 & Gemma 4 & 0.65$\pm$0.07 & 0.49$\pm$0.08 & 0.63$\pm$0.07 & 0.56$\pm$0.08 & \textbf{0.79$\pm$0.06} & \textbf{0.79$\pm$0.06} \\
\rowcolor{propbg}\cellcolor{white} & GPT-5.4 mini & \textbf{0.95$\pm$0.03} & \notrun & \notrun & \notrun & 0.86$\pm$0.05 & \notrun \\
\rowcolor{propbg}\cellcolor{white}\multirow{-6}{=}{Sneaky Sasquatch ($N=43$)} & Claude Haiku 4.5 & \textbf{0.88$\pm$0.05} & \notrun & \notrun & \notrun & 0.86$\pm$0.05 & \notrun \\
\midrule
 & Llama 3.1 & 0.23$\pm$0.06 & 0.37$\pm$0.06 & 0.30$\pm$0.06 & 0.26$\pm$0.06 & 0.56$\pm$0.07 & \textbf{0.58$\pm$0.07} \\
 & Qwen3 & 0.54$\pm$0.07 & 0.67$\pm$0.06 & 0.56$\pm$0.07 & 0.37$\pm$0.06 & 0.67$\pm$0.06 & \textbf{0.68$\pm$0.06} \\
 & Qwen3.6 & 0.68$\pm$0.06 & \textbf{0.77$\pm$0.06} & 0.60$\pm$0.07 & 0.30$\pm$0.06 & 0.63$\pm$0.06 & 0.65$\pm$0.06 \\
 & Gemma 4 & 0.61$\pm$0.07 & 0.65$\pm$0.06 & 0.60$\pm$0.07 & 0.37$\pm$0.06 & 0.58$\pm$0.07 & \textbf{0.67$\pm$0.06} \\
\rowcolor{propbg}\cellcolor{white} & GPT-5.4 mini & 0.61$\pm$0.07 & \notrun & \notrun & \notrun & \textbf{0.63$\pm$0.06} & \notrun \\
\rowcolor{propbg}\cellcolor{white}\multirow{-6}{=}{$\tau^2$-retail ($N=57$)} & Claude Haiku 4.5 & 0.42$\pm$0.07 & \notrun & \notrun & \notrun & \textbf{0.60$\pm$0.07} & \notrun \\
\midrule
 & Llama 3.1 & 0.30$\pm$0.07 & 0.57$\pm$0.08 & 0.43$\pm$0.08 & 0.36$\pm$0.07 & \textbf{0.70$\pm$0.07} & 0.68$\pm$0.07 \\
 & Qwen3 & 0.57$\pm$0.08 & 0.61$\pm$0.07 & 0.52$\pm$0.08 & 0.39$\pm$0.07 & \textbf{0.80$\pm$0.06} & 0.73$\pm$0.07 \\
 & Qwen3.6 & \textbf{0.75$\pm$0.07} & 0.70$\pm$0.07 & 0.61$\pm$0.07 & 0.39$\pm$0.07 & 0.73$\pm$0.07 & 0.68$\pm$0.07 \\
 & Gemma 4 & 0.61$\pm$0.07 & 0.61$\pm$0.07 & 0.59$\pm$0.07 & 0.36$\pm$0.07 & \textbf{0.68$\pm$0.07} & 0.59$\pm$0.07 \\
\rowcolor{propbg}\cellcolor{white} & GPT-5.4 mini & \textbf{0.73$\pm$0.07} & \notrun & \notrun & \notrun & 0.70$\pm$0.07 & \notrun \\
\rowcolor{propbg}\cellcolor{white}\multirow{-6}{=}{$\tau^2$-airline ($N=44$)} & Claude Haiku 4.5 & \textbf{0.75$\pm$0.07} & \notrun & \notrun & \notrun & 0.70$\pm$0.07 & \notrun \\
\bottomrule
\end{tabular}
\end{table}

\clearpage
\subsection{Evaluation on SOP-Bench}\label{app:sopbench}
In addtion to the MANTRA and the AgentOrca benchmark reported on in Table~\ref{tab:passsup5-by-domain} in Sec.~\ref{sec:experiments}, we also ran both shields on two SOP-Bench domains, patient intake and customer service, with the same four open-weight agents and settings.
While analysing the runs, we found that the benchmark's scorer does not read the arguments of tool calls and therefore accepts incorrect traces.
We first document this flaw and then report the results in Table~\ref{tab:sopbench-main}.

In the patient-intake domain, there are several tasks that require the agent to call some lookup tools, such as \texttt{validateInsurance}, and then call \texttt{registerPatient} with the results of these lookups as input.
Registration is successful if, among other conditions, the insurance status is \texttt{valid}.
However, the benchmark's provided mock implementation of the \texttt{registerPatient} tool ignores the input arguments it receives and always returns the (pre-stored in the database) correct output for the patient.
Further, the benchmark's scorer only checks the tool outputs, and not their arguments.
Together, these two issues can lead to incorrect traces being accepted as valid by the benchmark's scorer, when the agent provides incorrect inputs to the \texttt{registerPatient} tool (that were not the same as the results of the lookup tools), and then the mock implementation of \texttt{registerPatient} returns the correct output regardless of its inputs.
This was a common occurence in our experiments.
Agents often ignored the results of the lookups and provided incorrect inputs to the \texttt{registerPatient} tool, yet the benchmark's scorer still accepted these traces as valid due to the issues described above.
In particular, over 5 iterations on the 35 tasks in this domain, $68\%$ of "successful" runs by unshielded Qwen3, and $48\%$ of "successful" runs by unshielded Llama~3.1, contained a \texttt{registerPatient} call whose inputs did not match the results of the lookup tools.
These runs should be classified as non-compliant.
In contrast, with the prompt shield, none of the "successful" runs has this problem, because one of the shield's rules blocks any \texttt{registerPatient} call whose statuses differ from the lookup results.

Table~\ref{tab:sopbench-main} reports the results of our runs using SOP-Bench's flawed Task Success Rate (TSR) score.
We also observed, similarly to \mantra, that Llama 3.1 struggled to follow the prompt shield's advice and kept repeating blocked calls, but its performance improved under the token shield.

\begin{table}[htbp]
\centering
\setlength{\tabcolsep}{3pt}
\caption{Task Success Rate (TSR), SOP-Bench's flawed benchmark score, of open-weights agents on two domains.}
\label{tab:sopbench-main}
\begin{tabular}{@{}llrrr@{}}
\toprule
& & \multicolumn{3}{c}{TSR} \\
\cmidrule(l){3-5}
Domain & Model & None & \multicolumn{1}{c}{Ours} & \multicolumn{1}{c@{}}{Ours} \\
& & & \multicolumn{1}{c}{(prompt)} & \multicolumn{1}{c@{}}{(token)} \\
\midrule
\multirow{4}{*}{\shortstack[l]{Patient intake\\($N=35$)}} 
& Llama 3.1 & 0.03$\pm$0.05 & 0.29$\pm$0.28 & \textbf{0.38$\pm$0.20} \\
& Qwen3 & 0.29$\pm$0.11 & \textbf{0.61$\pm$0.04} & 0.46$\pm$0.07 \\
& Qwen3.6 & 0.90$\pm$0.07 & \textbf{1.00$\pm$0.00} & \textbf{1.00$\pm$0.00} \\
& Gemma 4 & \textbf{1.00$\pm$0.00} & \textbf{1.00$\pm$0.00} & \textbf{1.00$\pm$0.00} \\
\midrule
\multirow{4}{*}{\shortstack[l]{Customer service\\($N=35$)}} 
& Llama 3.1 & 0.10$\pm$0.07 & \textbf{0.19$\pm$0.08} & 0.18$\pm$0.03 \\
& Qwen3 & 0.46$\pm$0.06 & \textbf{0.63$\pm$0.03} & 0.59$\pm$0.04 \\
& Qwen3.6 & 0.90$\pm$0.08 & \textbf{0.91$\pm$0.05} & 0.90$\pm$0.05 \\
& Gemma 4 & 0.49$\pm$0.14 & \textbf{0.59$\pm$0.08} & 0.47$\pm$0.19 \\
\bottomrule
\end{tabular}
\end{table}

\clearpage
\subsection{Ablation Studies}\label{app:ablations}
We vary the three LLM-dependent parts of our framework one at a time.
Firstly, Table~\ref{tab:ablation-translator-mantra-passhatk} varies the model used in our autoformalization loop.

\begin{table}[htbp]
\centering
\small
\setlength{\tabcolsep}{4pt}
\caption{ \passSupThree{} of the prompt shield on two \mantra domains and three agents, with the specification written by different models.}
\label{tab:ablation-translator-mantra-passhatk}
\begin{tabular}{llrrrrr}
\toprule
Domain & Model & Claude Opus 5 & Claude Sonnet & gpt-5.5 & gpt-5.6-luna & Qwen3.8 \\
\midrule
\multirow{3}{*}{Operations ($N=47$)} & Llama 3.1 & 0.38$\pm$0.06 & 0.09$\pm$0.04 & \textbf{0.47$\pm$0.07} & 0.13$\pm$0.05 & 0.11$\pm$0.05 \\
 & Qwen3 & 0.64$\pm$0.07 & 0.34$\pm$0.07 & \textbf{0.68$\pm$0.07} & 0.23$\pm$0.06 & 0.34$\pm$0.07 \\
 & Gemma 4 & 0.75$\pm$0.06 & 0.68$\pm$0.07 & \textbf{0.79$\pm$0.06} & 0.36$\pm$0.07 & 0.55$\pm$0.07 \\
\midrule
\multirow{3}{*}{$\tau^2$-retail ($N=57$)} & Llama 3.1 & \textbf{0.28$\pm$0.05} & 0.25$\pm$0.06 & 0.09$\pm$0.04 & 0.16$\pm$0.05 & 0.14$\pm$0.05 \\
 & Qwen3 & 0.39$\pm$0.06 & \textbf{0.46$\pm$0.07} & 0.39$\pm$0.07 & 0.28$\pm$0.06 & 0.42$\pm$0.07 \\
 & Gemma 4 & 0.33$\pm$0.06 & 0.32$\pm$0.06 & \textbf{0.35$\pm$0.06} & 0.30$\pm$0.06 & \textbf{0.35$\pm$0.06} \\
\bottomrule
\end{tabular}
\end{table}

Table~\ref{tab:llmgen-passsupk} varies the autoformalization technique used to generate the rules for the \textsc{AgentSpec} benchmark.
\textsc{AgentSpec} provides their own language in which to write rules and suggests an in-context example generation approach.
We compare three approaches.
We use the suggested in-context example approach to write the rules in the provided language, using 4-5 testcases per domain as examples (AgentSpec, LLM-gen).
Then, we use our own autoformalization loop to generate the rules in the same language (AgentSpec, our loop).
Finally, we use our own autoformalization loop and specification (Ours).
All three autoformalization techniques used Claude Opus 5, and are evaluated on the same set of testcases (after removing the examples for AgentSpec, LLM-gen) with the prompt-shield.

\begin{table}[htbp]
\centering
\small
\setlength{\tabcolsep}{4pt}
\caption{\passSupFive{}~$\pm$~standard error over the $N$ held-out cases of a domain, for the prompt shield with \textsc{AgentSpec} rules written by \textsc{AgentSpec}'s LLM rule generator (LLM-gen) and by our autoformalization loop (our loop), next to the unshielded agent (None) and our specification (Ours).}
\begin{tabular}{llrrrr}
\toprule
Domain & Model & None & AgentSpec (LLM-gen) & AgentSpec (our loop) & Ours \\
\midrule
\multirow{4}{*}{\shortstack[l]{Operations\\($N=43$)}} & Llama 3.1 & 0.00$\pm$0.00 & 0.05$\pm$0.03 & 0.00$\pm$0.00 & \textbf{0.26$\pm$0.07} \\
 & Qwen3 & 0.28$\pm$0.07 & 0.26$\pm$0.07 & 0.19$\pm$0.06 & \textbf{0.58$\pm$0.08} \\
 & Qwen3.6 & 0.60$\pm$0.08 & \textbf{0.77$\pm$0.07} & \textbf{0.77$\pm$0.07} & 0.70$\pm$0.07 \\
 & Gemma 4 & 0.56$\pm$0.08 & 0.63$\pm$0.07 & 0.56$\pm$0.08 & \textbf{0.70$\pm$0.07} \\
\midrule
\multirow{4}{*}{\shortstack[l]{Sneaky Sasquatch\\($N=39$)}} & Llama 3.1 & 0.08$\pm$0.04 & 0.03$\pm$0.03 & 0.00$\pm$0.00 & \textbf{0.15$\pm$0.06} \\
 & Qwen3 & 0.44$\pm$0.08 & 0.23$\pm$0.07 & 0.41$\pm$0.08 & \textbf{0.69$\pm$0.07} \\
 & Qwen3.6 & 0.54$\pm$0.08 & 0.51$\pm$0.08 & 0.51$\pm$0.08 & \textbf{0.62$\pm$0.08} \\
 & Gemma 4 & 0.44$\pm$0.08 & 0.26$\pm$0.07 & 0.41$\pm$0.08 & \textbf{0.56$\pm$0.08} \\
\midrule
\multirow{4}{*}{\shortstack[l]{$\tau^2$-retail\\($N=52$)}} & Llama 3.1 & 0.04$\pm$0.03 & 0.08$\pm$0.04 & 0.08$\pm$0.04 & \textbf{0.19$\pm$0.06} \\
 & Qwen3 & 0.15$\pm$0.05 & 0.27$\pm$0.06 & \textbf{0.33$\pm$0.07} & 0.31$\pm$0.06 \\
 & Qwen3.6 & 0.33$\pm$0.07 & 0.38$\pm$0.07 & 0.31$\pm$0.06 & \textbf{0.40$\pm$0.07} \\
 & Gemma 4 & 0.15$\pm$0.05 & \textbf{0.25$\pm$0.06} & 0.15$\pm$0.05 & \textbf{0.25$\pm$0.06} \\
\midrule
\multirow{4}{*}{\shortstack[l]{$\tau^2$-airline\\($N=40$)}} & Llama 3.1 & 0.33$\pm$0.07 & 0.35$\pm$0.08 & 0.33$\pm$0.07 & \textbf{0.40$\pm$0.08} \\
 & Qwen3 & 0.30$\pm$0.07 & 0.28$\pm$0.07 & 0.35$\pm$0.08 & \textbf{0.55$\pm$0.08} \\
 & Qwen3.6 & 0.47$\pm$0.08 & 0.35$\pm$0.08 & 0.42$\pm$0.08 & \textbf{0.57$\pm$0.08} \\
 & Gemma 4 & 0.42$\pm$0.08 & 0.35$\pm$0.08 & 0.45$\pm$0.08 & \textbf{0.60$\pm$0.08} \\
\bottomrule
\end{tabular}
\label{tab:llmgen-passsupk}
\end{table}

Finally, Table~\ref{tab:ablation-labeler-mantra-passhatk} modifies the LLM used for labeling, including our finetuned Qwen3-4B labeler decsribed in App.~\ref{app:finetune-labeler}.
Table~\ref{tab:labeler-passsup5} shows a more detailed comparison of GPT-4.1 and GPT-4.1-mini as labelers on \mantra.

\begin{table}[htbp]
\centering
\footnotesize
\setlength{\tabcolsep}{3pt}
\caption{Labeler ablation: \passSupThree{} of the prompt shield on two \mantra domains and two agents.
}
\label{tab:ablation-labeler-mantra-passhatk}
\begin{tabular}{lrrrr}
\toprule
& \multicolumn{4}{c}{\passSupThree} \\
\cmidrule(lr){2-5}
& \multicolumn{2}{c}{Sneaky Sasquatch} & \multicolumn{2}{c}{$\tau^2$-retail} \\
\cmidrule(lr){2-3} \cmidrule(lr){4-5}
Labeler & Qwen3 & Gemma 4 & Qwen3 & Gemma 4 \\
\midrule
gpt-4.1-mini & 0.70$\pm$0.07 & 0.62$\pm$0.07 & 0.39$\pm$0.06 & \textbf{0.33$\pm$0.06} \\
Qwen3-4B & 0.68$\pm$0.07 & 0.60$\pm$0.07 & \textbf{0.44$\pm$0.06} & 0.28$\pm$0.05 \\
Gemma 4 & 0.07$\pm$0.04 & 0.07$\pm$0.04 & 0.42$\pm$0.07 & 0.28$\pm$0.06 \\
gpt-oss & 0.02$\pm$0.02 & 0.07$\pm$0.04 & 0.35$\pm$0.06 & 0.26$\pm$0.06 \\
Llama 3.1 & 0.02$\pm$0.02 & 0.05$\pm$0.03 & \textbf{0.44$\pm$0.07} & 0.28$\pm$0.06 \\
Qwen3 & 0.05$\pm$0.03 & 0.07$\pm$0.04 & 0.35$\pm$0.06 & 0.26$\pm$0.06 \\
Qwen3.6 & 0.02$\pm$0.02 & 0.09$\pm$0.04 & 0.40$\pm$0.07 & 0.25$\pm$0.06 \\
gpt-4.1 & \textbf{0.80$\pm$0.06} & \textbf{0.79$\pm$0.06} & 0.35$\pm$0.06 & 0.31$\pm$0.06 \\
\bottomrule
\end{tabular}
\end{table}

\begin{table}[htbp]
\centering
\small
\setlength{\tabcolsep}{4pt}
\caption{Labeler ablation: \passSupFive{} on all \mantra with the GPT-4.1 mini and the GPT-4.1 labeler.}
\begin{tabular}{llrrr}
\toprule
& & \multicolumn{1}{c}{Prompt shield} & \multicolumn{1}{c}{Prompt shield} & \multicolumn{1}{c}{Token shield} \\
& & \multicolumn{1}{c}{(gpt-4.1-mini)} & \multicolumn{1}{c}{(gpt-4.1)} & \multicolumn{1}{c}{(gpt-4.1)} \\
\midrule
Domain & Model & \passSupK{5} & \passSupK{5} & \passSupK{5} \\
\midrule
\multirow{4}{*}{Operations} & Llama 3.1 & 0.28$\pm$0.07 & 0.21$\pm$0.06 & \textbf{0.38$\pm$0.07} \\
 & Qwen3 & 0.62$\pm$0.07 & \textbf{0.64$\pm$0.07} & 0.55$\pm$0.07 \\
 & Qwen3.6 & \textbf{0.72$\pm$0.07} & 0.70$\pm$0.07 & 0.66$\pm$0.07 \\
 & Gemma 4 & 0.72$\pm$0.07 & \textbf{0.77$\pm$0.06} & 0.57$\pm$0.07 \\
\midrule
\multirow{4}{*}{Sneaky Sasquatch} & Llama 3.1 & 0.14$\pm$0.05 & 0.09$\pm$0.04 & \textbf{0.42$\pm$0.08} \\
 & Qwen3 & 0.67$\pm$0.07 & \textbf{0.77$\pm$0.07} & 0.37$\pm$0.07 \\
 & Qwen3.6 & 0.63$\pm$0.07 & \textbf{0.65$\pm$0.07} & 0.58$\pm$0.08 \\
 & Gemma 4 & 0.58$\pm$0.08 & \textbf{0.79$\pm$0.06} & 0.65$\pm$0.07 \\
\midrule
\multirow{4}{*}{$\tau^2$-retail} & Llama 3.1 & 0.21$\pm$0.05 & 0.18$\pm$0.05 & \textbf{0.30$\pm$0.06} \\
 & Qwen3 & 0.32$\pm$0.06 & 0.26$\pm$0.06 & \textbf{0.40$\pm$0.07} \\
 & Qwen3.6 & \textbf{0.42$\pm$0.07} & 0.39$\pm$0.07 & 0.18$\pm$0.05 \\
 & Gemma 4 & \textbf{0.26$\pm$0.06} & 0.25$\pm$0.06 & 0.25$\pm$0.06 \\
\midrule
\multirow{4}{*}{$\tau^2$-airline} & Llama 3.1 & 0.41$\pm$0.07 & 0.36$\pm$0.07 & \textbf{0.57$\pm$0.08} \\
 & Qwen3 & 0.59$\pm$0.07 & 0.57$\pm$0.08 & \textbf{0.61$\pm$0.07} \\
 & Qwen3.6 & 0.61$\pm$0.07 & \textbf{0.64$\pm$0.07} & 0.55$\pm$0.08 \\
 & Gemma 4 & 0.64$\pm$0.07 & \textbf{0.66$\pm$0.07} & 0.50$\pm$0.08 \\
\bottomrule
\end{tabular}
\label{tab:labeler-passsup5}
\end{table}

\clearpage

\subsection{Labeler Finetuning Details}\label{app:finetune-labeler}

We finetune Qwen3-4B-Instruct-2507 to act as a labeler, via distillation from labels produced by GPT-4.1.

\paragraph{Task and data.} 
Each training example is one labeler call exactly as issued by the shield as described in Sec.~\ref{sec:online}.
We use the same input that the GPT-4.1 labeler saw, and the target output is GPT-4.1's label.
As training data, we used three sources.
Firstly, we used existing GPT-4.1 labels from the runs in Sec.~\ref{sec:experiments}.
Specifically, 239 examples from the Operations and Sneaky Sasquatch domains from \mantra, 6408 from the $\tau^2$ domains from \mantra and 70 from SOP-Bench.
Secondly, we paraphrased 3 times, using Codex, 160 examples from the real runs, asked GPT-4.1 to label the paraphrased examples, and kept 441 examples whose labels matched the original labels.
Lastly, we generated 48 new synthetic domains using Codex, each with a specification, set of texts for labeling, and intended label.
We used GPT-4.1 to label these texts and kept 3133 examples whose labels matched the intended labels.

\paragraph{Training and evaluation.} We train using a leave-one-benchmark-family-out procedure over the the two non-$\tau^2$ \mantra domains, the two $\tau^2$ \mantra domains, and SOP-Bench.
The left out family is used for the evaluation reported in Table~\ref{tab:shield-comparison-passsup5}, while the other two are used for training and validation (10\%).
We fine-tune using LoRA for 2 epochs with learning rate \(10^{-4}\) and effective batch size 16. The loss is computed only over target tokens and thinking mode is disabled. We select the checkpoint with lowest validation loss. 
For evaluation, we use the prompt shield over 5 runs with 4 agents as described in Sec.~\ref{sec:experiments}, and in Table~\ref{tab:shield-comparison-passsup5} show that our fine-tuned labeler is comparable with the original GPT-4.1 labeler.

\begin{table}[h]
\centering
\footnotesize
\setlength{\tabcolsep}{3pt}
\caption{\passSupFive{}~$\pm$~standard error over the $N$ test cases of a domain, on the \mantra synthetic domains (top) and the \mantra $\tau^2$ domains (bottom), without a shield (None) and with our prompt shield, with gpt-4.1 or the fine-tuned Qwen3-4B as the labeler.}
\label{tab:shield-comparison-passsup5}
\begin{tabular}{llrrr}
\toprule
& & \multicolumn{3}{c}{\passSupFive} \\
\cmidrule(lr){3-5}
Domain & Model & None & Ours (gpt-4.1) & Ours (Qwen3-4B FT) \\
\midrule
\multirow{4}{*}{\shortstack[l]{Operations\\($N=47$)}} & Llama 3.1 & 0.00$\pm$0.00 & 0.21$\pm$0.06 & \textbf{0.30$\pm$0.07} \\
 & Qwen3 & 0.28$\pm$0.07 & \textbf{0.64$\pm$0.07} & 0.60$\pm$0.07 \\
 & Qwen3.6 & 0.62$\pm$0.07 & \textbf{0.72$\pm$0.07} & \textbf{0.72$\pm$0.07} \\
 & Gemma 4 & 0.55$\pm$0.07 & \textbf{0.77$\pm$0.06} & 0.70$\pm$0.07 \\
\midrule
\multirow{4}{*}{\shortstack[l]{Sneaky Sasquatch\\($N=43$)}} & Llama 3.1 & 0.07$\pm$0.04 & 0.09$\pm$0.04 & \textbf{0.14$\pm$0.05} \\
 & Qwen3 & 0.44$\pm$0.08 & \textbf{0.77$\pm$0.07} & 0.65$\pm$0.07 \\
 & Qwen3.6 & 0.56$\pm$0.08 & \textbf{0.65$\pm$0.07} & 0.58$\pm$0.08 \\
 & Gemma 4 & 0.42$\pm$0.08 & \textbf{0.79$\pm$0.06} & 0.58$\pm$0.08 \\
\cmidrule[\lightrulewidth](l{0pt}r{0pt}){1-5}\morecmidrules
\cmidrule[\lightrulewidth](l{0pt}r{0pt}){1-5}
\multirow{4}{*}{\shortstack[l]{$\tau^2$-retail\\($N=57$)}} & Llama 3.1 & 0.05$\pm$0.03 & 0.18$\pm$0.05 & \textbf{0.28$\pm$0.06} \\
 & Qwen3 & 0.16$\pm$0.05 & 0.28$\pm$0.06 & \textbf{0.37$\pm$0.06} \\
 & Qwen3.6 & 0.39$\pm$0.07 & 0.40$\pm$0.07 & \textbf{0.46$\pm$0.07} \\
 & Gemma 4 & 0.18$\pm$0.05 & \textbf{0.25$\pm$0.06} & 0.23$\pm$0.06 \\
\midrule
\multirow{4}{*}{\shortstack[l]{$\tau^2$-airline\\($N=44$)}} & Llama 3.1 & 0.34$\pm$0.07 & 0.36$\pm$0.07 & \textbf{0.43$\pm$0.08} \\
 & Qwen3 & 0.32$\pm$0.07 & 0.57$\pm$0.08 & \textbf{0.66$\pm$0.07} \\
 & Qwen3.6 & 0.50$\pm$0.08 & \textbf{0.64$\pm$0.07} & \textbf{0.64$\pm$0.07} \\
 & Gemma 4 & 0.41$\pm$0.07 & \textbf{0.66$\pm$0.07} & 0.64$\pm$0.07 \\
\bottomrule
\end{tabular}
\end{table}

\clearpage
\subsection{Cost and Latency}\label{app:cost-latency}
We report the seconds per tool call in Table~\ref{tab:latency-per-call}, the number of proposed tool calls per case in Table~\ref{tab:calls-per-case}, and the labeling cost in API calls in Table~\ref{tab:ours-labeler-cost-short} for the \mantra benchmark.

\begin{table}[h]
\centering
\scriptsize
\setlength{\tabcolsep}{3pt}
\caption{Seconds per proposed tool call (including blocked), mean $\pm$ standard deviation.}
\begin{tabular}{llrrrrrr}
\toprule
Domain & Model & None & AgentSpec & Progent & DynaGuard & Ours (prompt) & Ours (token) \\
\midrule
\multirow{4}{*}{Operations ($N=47$)} & Llama 3.1 & 0.59 $\pm$ 0.03 & 0.57 $\pm$ 0.02 & 30.70 $\pm$ 0.46 & 3.01 $\pm$ 0.07 & 0.73 $\pm$ 0.04 & 2.26 $\pm$ 0.05 \\
 & Qwen3 & 10.00 $\pm$ 0.67 & 9.95 $\pm$ 0.13 & 44.31 $\pm$ 1.29 & 13.28 $\pm$ 0.20 & 9.58 $\pm$ 0.55 & 10.52 $\pm$ 0.18 \\
 & Qwen3.6 & 18.96 $\pm$ 0.82 & 25.72 $\pm$ 2.31 & 54.21 $\pm$ 1.85 & 65.25 $\pm$ 5.64 & 29.12 $\pm$ 2.98 & 23.87 $\pm$ 2.90 \\
 & Gemma 4 & 0.64 $\pm$ 0.01 & 0.64 $\pm$ 0.01 & 34.47 $\pm$ 0.92 & 2.63 $\pm$ 0.06 & 0.75 $\pm$ 0.03 & 0.99 $\pm$ 0.03 \\
\midrule
\multirow{4}{*}{Sneaky Sasquatch ($N=43$)} & Llama 3.1 & 0.39 $\pm$ 0.01 & 0.40 $\pm$ 0.01 & 21.25 $\pm$ 0.98 & 2.34 $\pm$ 0.14 & 0.45 $\pm$ 0.00 & 1.66 $\pm$ 0.04 \\
 & Qwen3 & 7.52 $\pm$ 0.69 & 8.89 $\pm$ 0.28 & 28.77 $\pm$ 1.08 & 10.68 $\pm$ 0.79 & 6.72 $\pm$ 0.22 & 10.57 $\pm$ 1.47 \\
 & Qwen3.6 & 15.80 $\pm$ 1.40 & 20.24 $\pm$ 2.36 & 41.30 $\pm$ 1.50 & 29.24 $\pm$ 2.55 & 13.61 $\pm$ 0.55 & 12.95 $\pm$ 1.53 \\
 & Gemma 4 & 0.45 $\pm$ 0.01 & 0.46 $\pm$ 0.01 & 25.33 $\pm$ 1.12 & 2.36 $\pm$ 0.06 & 0.48 $\pm$ 0.02 & 0.65 $\pm$ 0.03 \\
\midrule
\multirow{4}{*}{$\tau^2$-retail ($N=57$)} & Llama 3.1 & 0.51 $\pm$ 0.02 & 0.50 $\pm$ 0.01 & 21.75 $\pm$ 2.63 & 3.24 $\pm$ 0.07 & 0.80 $\pm$ 0.02 & 2.18 $\pm$ 0.04 \\
 & Qwen3 & 16.03 $\pm$ 1.44 & 14.39 $\pm$ 0.94 & 39.36 $\pm$ 1.66 & 19.02 $\pm$ 0.83 & 13.03 $\pm$ 0.27 & 14.77 $\pm$ 1.25 \\
 & Qwen3.6 & 53.32 $\pm$ 4.48 & 51.84 $\pm$ 2.93 & 79.09 $\pm$ 2.69 & 42.76 $\pm$ 2.30 & 47.04 $\pm$ 2.90 & 39.88 $\pm$ 2.87 \\
 & Gemma 4 & 0.75 $\pm$ 0.03 & 0.77 $\pm$ 0.04 & 31.55 $\pm$ 1.47 & 2.75 $\pm$ 0.08 & 0.92 $\pm$ 0.08 & 1.27 $\pm$ 0.07 \\
\midrule
\multirow{4}{*}{$\tau^2$-airline ($N=44$)} & Llama 3.1 & 2.12 $\pm$ 0.09 & 1.07 $\pm$ 0.11 & 12.95 $\pm$ 2.22 & 3.46 $\pm$ 0.39 & 1.00 $\pm$ 0.07 & 1.82 $\pm$ 0.02 \\
 & Qwen3 & 18.79 $\pm$ 0.71 & 20.77 $\pm$ 0.83 & 48.00 $\pm$ 2.88 & 25.87 $\pm$ 1.67 & 16.58 $\pm$ 0.43 & 14.40 $\pm$ 1.28 \\
 & Qwen3.6 & 50.06 $\pm$ 2.11 & 45.92 $\pm$ 3.36 & 86.03 $\pm$ 7.12 & 46.95 $\pm$ 3.85 & 49.30 $\pm$ 6.83 & 36.78 $\pm$ 3.93 \\
 & Gemma 4 & 1.44 $\pm$ 0.03 & 1.33 $\pm$ 0.08 & 45.91 $\pm$ 1.35 & 4.01 $\pm$ 0.14 & 1.35 $\pm$ 0.08 & 0.98 $\pm$ 0.05 \\
\bottomrule
\end{tabular}
\label{tab:latency-per-call}
\end{table}

\begin{table}[h]
\centering
\scriptsize
\setlength{\tabcolsep}{3pt}
\caption{Number of tool calls the agent proposed per testcase (including blocked) mean $\pm$ standard deviation.}
\begin{tabular}{llrrrrrr}
\toprule
Domain & Model & None & AgentSpec & Progent & DynaGuard & Ours (prompt) & Ours (token) \\
\midrule
\multirow{4}{*}{Operations ($N=47$)} & Llama 3.1 & 4.8 $\pm$ 0.1 & 7.4 $\pm$ 0.3 & 5.0 $\pm$ 0.2 & 18.1 $\pm$ 1.1 & 8.7 $\pm$ 0.5 & 11.0 $\pm$ 0.3 \\
 & Qwen3 & 4.3 $\pm$ 0.1 & 4.9 $\pm$ 0.2 & 4.4 $\pm$ 0.1 & 13.3 $\pm$ 0.9 & 5.3 $\pm$ 0.1 & 8.3 $\pm$ 0.9 \\
 & Qwen3.6 & 5.0 $\pm$ 0.1 & 5.3 $\pm$ 0.2 & 5.1 $\pm$ 0.3 & 9.0 $\pm$ 0.8 & 6.3 $\pm$ 0.3 & 10.9 $\pm$ 0.4 \\
 & Gemma 4 & 5.2 $\pm$ 0.1 & 5.2 $\pm$ 0.0 & 5.1 $\pm$ 0.1 & 5.3 $\pm$ 0.2 & 6.1 $\pm$ 0.6 & 10.5 $\pm$ 1.0 \\
\midrule
\multirow{4}{*}{Sneaky Sasquatch ($N=43$)} & Llama 3.1 & 8.6 $\pm$ 1.5 & 11.8 $\pm$ 1.1 & 8.7 $\pm$ 1.1 & 10.8 $\pm$ 1.1 & 9.0 $\pm$ 0.3 & 8.8 $\pm$ 0.6 \\
 & Qwen3 & 4.2 $\pm$ 0.2 & 5.6 $\pm$ 0.3 & 4.5 $\pm$ 0.2 & 6.1 $\pm$ 0.6 & 5.6 $\pm$ 0.1 & 6.5 $\pm$ 0.3 \\
 & Qwen3.6 & 4.5 $\pm$ 0.1 & 6.6 $\pm$ 0.1 & 4.7 $\pm$ 0.3 & 5.4 $\pm$ 0.2 & 5.6 $\pm$ 0.1 & 6.2 $\pm$ 0.6 \\
 & Gemma 4 & 4.3 $\pm$ 0.4 & 4.3 $\pm$ 0.2 & 4.5 $\pm$ 0.4 & 4.6 $\pm$ 0.6 & 7.0 $\pm$ 0.2 & 7.4 $\pm$ 0.5 \\
\midrule
\multirow{4}{*}{$\tau^2$-retail ($N=57$)} & Llama 3.1 & 10.1 $\pm$ 0.9 & 9.7 $\pm$ 0.7 & 10.4 $\pm$ 0.7 & 12.4 $\pm$ 1.7 & 8.2 $\pm$ 1.0 & 23.2 $\pm$ 1.7 \\
 & Qwen3 & 3.7 $\pm$ 0.2 & 3.8 $\pm$ 0.1 & 3.8 $\pm$ 0.1 & 4.6 $\pm$ 0.1 & 6.2 $\pm$ 0.2 & 12.4 $\pm$ 2.1 \\
 & Qwen3.6 & 5.9 $\pm$ 0.3 & 6.0 $\pm$ 0.2 & 5.6 $\pm$ 0.2 & 11.3 $\pm$ 1.0 & 10.1 $\pm$ 1.4 & 15.4 $\pm$ 1.0 \\
 & Gemma 4 & 2.6 $\pm$ 0.2 & 2.7 $\pm$ 0.2 & 2.7 $\pm$ 0.2 & 2.3 $\pm$ 0.1 & 5.2 $\pm$ 0.2 & 9.1 $\pm$ 1.6 \\
\midrule
\multirow{4}{*}{$\tau^2$-airline ($N=44$)} & Llama 3.1 & 15.3 $\pm$ 2.6 & 6.7 $\pm$ 0.6 & 5.5 $\pm$ 0.9 & 4.0 $\pm$ 0.3 & 6.3 $\pm$ 0.7 & 27.9 $\pm$ 0.7 \\
 & Qwen3 & 3.1 $\pm$ 0.5 & 2.5 $\pm$ 0.1 & 2.9 $\pm$ 0.3 & 1.8 $\pm$ 0.1 & 4.2 $\pm$ 0.3 & 14.2 $\pm$ 1.2 \\
 & Qwen3.6 & 4.5 $\pm$ 0.4 & 3.9 $\pm$ 0.6 & 4.1 $\pm$ 0.3 & 4.9 $\pm$ 0.4 & 5.0 $\pm$ 0.7 & 8.7 $\pm$ 1.1 \\
 & Gemma 4 & 1.8 $\pm$ 0.2 & 1.8 $\pm$ 0.2 & 1.8 $\pm$ 0.1 & 1.1 $\pm$ 0.1 & 2.2 $\pm$ 0.2 & 6.1 $\pm$ 0.7 \\
\bottomrule
\end{tabular}
\label{tab:calls-per-case}
\end{table}

\begin{table}[htbp]
\centering
\footnotesize
\setlength{\tabcolsep}{3pt}
\caption{Cost of the labeler per domain, in US cents per testcase and in USD per one run over the full domain, at prices of GPT-4.1 mini (0.40/1.60 USD per 1M input/output tokens) and GPT-4.1 (2.00/8.00).}
\label{tab:ours-labeler-cost-short}
\begin{tabular}{lrrrrr}
\toprule
& & \multicolumn{2}{c}{GPT-4.1 mini} & \multicolumn{2}{c}{GPT-4.1} \\
\cmidrule(lr){3-4} \cmidrule(lr){5-6}
Domain & cases & cents/case & USD/domain & cents/case & USD/domain \\
\midrule
Operations & 47 & 0.06 & 0.029 & 0.30 & 0.143 \\
Sneaky Sasquatch & 43 & 0.07 & 0.030 & 0.35 & 0.150 \\
$\tau^2$-retail & 57 & 0.25 & 0.141 & 1.24 & 0.705 \\
$\tau^2$-airline & 44 & 0.20 & 0.086 & 0.97 & 0.429 \\
\bottomrule
\end{tabular}
\end{table}

\clearpage
\subsection{Case Study: Prompt Shield versus Token Shield}\label{app:prompt-vs-token}

Here, we show an example of a task from the Sneaky-Sasquatch domain of \mantra, in which Llama~3.1~8B does not follow the instructions of the prompt shield but succeeds with the token shield.

\paragraph{Task and Checks.}
The task is to treat a patient with low magnesium. 
The intended treatment involves listening to the patient, checking the inventory for food items, and giving an appropriate food to the patient.
Importantly, the checks for this testcase mandate that the call \texttt{check\_inventory(item\_type="foods")} must be made before the call \texttt{give\_food} with a suitable food argument (Milk, Relish or Cheese).

\paragraph{Prompt shield.}
Without a shield, the agent gives Milk without checking the inventory, which fails every time.
On the other hand, the prompt shield intervenes and blocks the \code{give\_food} call, suggesting the prerequisite \code{check\_inventory} call:
\begin{quote}\small\ttfamily
[SHIELD] Blocked give\_food(patient\_id="PT-6620", food\_name="Milk").\\
- Policy rule for patient PT-6620: \dots\ check what the inventory holds before you hand a food over.\\
\hspace*{1em}First: check\_inventory(item\_type="foods")
\end{quote}
However, the agent does not follow this advice.
Instead, it makes other calls such as \texttt{listen\_to\_patient} (allowed), \texttt{talk\_to\_head\_doctor}, (blocked), and repeats the blocked \texttt{give\_food} call.
The run ends without checking the inventory or giving food to the patient, so the patient remains untreated.

\paragraph{Token shield.}
The token shield masks the tokens for the \texttt{give\_food} call so the agent never generates it.
Instead, the token shield boosts the calls
\texttt{listen\_to\_patient} and \texttt{check\_inventory}, which are both called by the agent.
After the \texttt{check\_inventory} call, the environment reports that Cheese is available and the agent proceeds to give it to the patient, who is then treated successfully.

\label{endofdocument}
\newoutputstream{pagestotal}
\openoutputfile{main.pgt}{pagestotal}
\addtostream{pagestotal}{\getpagerefnumber{endofdocument}}
\closeoutputstream{pagestotal}

\end{document}